\documentclass{article}

\usepackage[final, nonatbib]{neurips_2022}

\usepackage[utf8]{inputenc} 
\usepackage[T1]{fontenc}    
\usepackage{hyperref}       
\usepackage{url}            
\usepackage{booktabs}       
\usepackage{amsfonts}       
\usepackage{nicefrac}       
\usepackage{microtype}      
\usepackage{xcolor}         
\usepackage[numbers]{natbib} 

\usepackage{amsmath}

\usepackage{amssymb}
\usepackage{bm}
\usepackage{cuted}
\usepackage{booktabs}
\usepackage{wrapfig}
\usepackage{xcolor}
\usepackage{graphicx}
\usepackage{multirow}
\usepackage{wrapfig}
\usepackage{caption}
\usepackage{subcaption}
\usepackage{tablefootnote}
\usepackage{comment}

\definecolor{myGray}{rgb}{0.9,0.9,0.9}
\newcommand{\bgray}[1]{\colorbox{myGray}{#1}}

\def\0{{\bf 0}}
\def\1{{\bf 1}}

\newcommand{\tabref}[1]{Table~\ref{#1}}
\newcommand{\secref}[1]{Sec.~\ref{#1}}
\newcommand{\figref}[1]{Fig.~\ref{#1}}

\newcommand{\eqnref}[1]{Eqn.~\ref{#1}}

\renewcommand{\paragraph}[1]{\textbf{#1. }}

\def\nbody{$N$-body MNIST}
\def\Cnino{\bm{C}^{\mathtt{Nino3.4}}}
\def\Cninok{C_k^{\mathtt{Nino3.4}}}

\long\def\comment#1{}

\title{Earthformer: Exploring Space-Time Transformers for Earth System Forecasting}

\author{%
  Zhihan Gao\thanks{Work done while being an intern at Amazon Web Services. $^\dagger$Contact person.} \\
  Hong Kong University of Science and Technology\\
  \texttt{zhihan.gao@connect.ust.hk} \\
  \And
  Xingjian Shi$^\dagger$\\
  Amazon Web Services \\
  \texttt{xjshi@amazon.com} \\
  \And
  Hao Wang \\
  Rutgers University\\
  \texttt{hw488@cs.rutgers.edu}\\
  \And
  Yi Zhu\\
  Amazon Web Services \\
  \texttt{yzaws@amazon.com} \\
  \And
  Yuyang Wang\\
  Amazon Web Services \\
  \texttt{yuyawang@amazon.com} \\
  \And
  Mu Li\\
  Amazon Web Services \\
  \texttt{mli@amazon.com} \\
  \And
  Dit-Yan Yeung\\
  Hong Kong University of Science and Technology \\
  \texttt{dyyeung@cse.ust.hk} \\
}

\begin{document}

\maketitle

\begin{abstract}
Conventionally, Earth system (e.g., weather and climate) forecasting relies on numerical simulation with complex physical models and hence is both expensive in computation and demanding on domain expertise. 
With the explosive growth of spatiotemporal Earth observation data in the past decade, data-driven models that apply Deep Learning (DL) are demonstrating impressive potential for various Earth system forecasting tasks.
The Transformer as an emerging DL architecture, despite its broad success in other domains, has limited adoption in this area.
In this paper, we propose \emph{Earthformer}, a space-time Transformer for Earth system forecasting. Earthformer is based on a generic, flexible and efficient space-time attention block, named \emph{Cuboid Attention}. The idea is to decompose the data into cuboids and apply cuboid-level self-attention in parallel. These cuboids are further connected with a collection of global vectors. We conduct experiments on the MovingMNIST dataset and a newly proposed chaotic \nbody{} dataset to verify the effectiveness of cuboid attention and figure out the best design of Earthformer. Experiments on two real-world benchmarks about precipitation nowcasting and El Niño/Southern Oscillation (ENSO) forecasting show that Earthformer achieves state-of-the-art performance.
\end{abstract}

\section{Introduction}

The Earth is a complex system. Variabilities of the Earth system, ranging from regular events like temperature fluctuation to extreme events like drought, hail storm, and El Niño/Southern Oscillation (ENSO), impact our daily life. Among all the consequences, Earth system variabilities can influence crop yields, delay airlines, cause floods and forest fires. Precise and timely forecasting of these variabilities can help people take necessary precautions to avoid crisis, or better utilize natural resources such as wind and solar energy. Thus, improving forecasting models for Earth variabilities~(e.g., weather and climate) has a huge socioeconomic impact. Despite its importance, the operational weather and climate forecasting systems have not fundamentally changed for almost 50 years~\cite{reichstein2019deep}. These operational models, including the state-of-the-art High Resolution Ensemble Forecast (HREF) rainfall nowcasting model used in National Oceanic and Atmospheric Administration (NOAA)~\cite{ravuri2021skilful}, rely on meticulous numerical simulation of physical models. Such simulation-based systems inevitably fall short in the ability to incorporate signals from newly emerging geophysical observation systems~\cite{goodman2019goes}, or take advantage of the Petabytes-scale Earth observation data~\cite{veillette2020sevir}.

As an appealing alternative, 
deep learning (DL) is offering a new approach for Earth system forecasting~\cite{reichstein2019deep}. Instead of explicitly incorporating physical rules, DL-based forecasting models are trained on the Earth observation data~\cite{shi2015convolutional}. 
By learning from a large amount of observations, DL models are able to figure out the system's intrinsic physical rules and generate predictions that outperform simulation-based models~\cite{espeholt2021skillful}. 
Such technique has demonstrated success in several applications, including precipitation nowcasting~\cite{ravuri2021skilful,de2020rainbench} and ENSO forecasting~\cite{ham2019deep}.
Because the Earth system is chaotic~\cite{letellier2019chaos}, high-dimensional, and spatiotemporal, designing appropriate DL architecture for modeling the system is particularly challenging. Previous works relied on the combination of Recurrent Neural Networks (RNN) and Convolutional Neural Networks (CNN)~\cite{shi2015convolutional, shi2017deep, veillette2020sevir, guen2020disentangling, wang2022predrnn}. These two architectures impose temporal and spatial inductive biases that help capturing spatiotemporal patterns. However, as a chaotic system, variabilities of the Earth system, such as rainfall and ENSO, are highly sensitive to the system's initial conditions and can respond abruptly to internal changes. It is unclear whether the inductive biases in RNN and CNN models still hold for such complex systems.




\begin{figure}[!tb]
    \centering
    \vskip -0.8cm
    \includegraphics[width=0.9\textwidth]{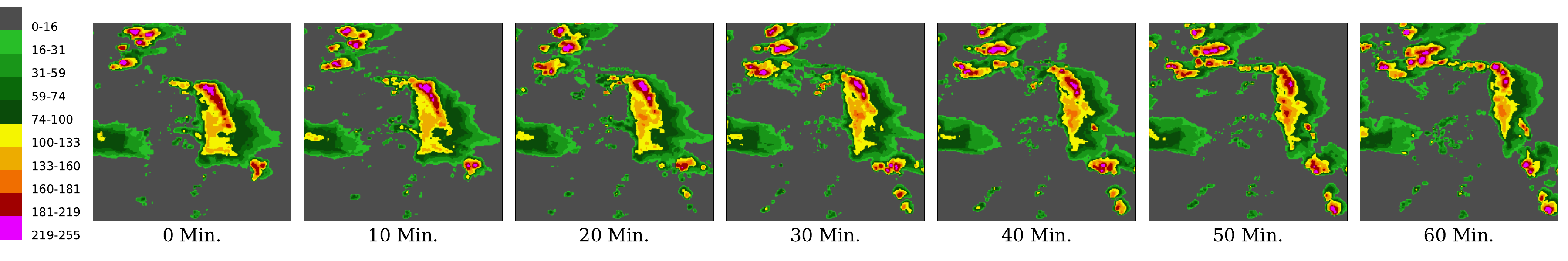}
    \vskip -0.2cm
    \caption{Example Vertically Integrated Liquid (VIL) observation sequence from the Storm EVent ImageRy (SEVIR) dataset. The observation intensity is mapped to pixel value of the range 0-255. The larger value indicates the higher precipitation intensity.}
    \label{fig:sevir_example}
    \vskip -0.6cm
\end{figure}

On the other hand, recent years have witnessed major breakthroughs in DL brought by the wide adoption of Transformer. The model was originally proposed for natural language processing~\cite{vaswani2017attention, devlin2018bert}, and later has been extended to computer vision~\cite{dosovitskiy2020image,liu2021swin}, multimodal text-image generation~\cite{ramesh2021zero}, graph learning~\cite{ying2021transformers}, etc. 
Transformer relies on the attention mechanism to capture data correlations and is powerful at modeling complex and long-range dependencies, both of which appear in Earth systems (See \figref{fig:sevir_example} for an example of Earth observation data). 
Despite being suitable for the problem, Transformer sees limited adoption for Earth system forecasting. 
Naively applying the Transformer architecture is infeasible because the $O(N^2)$ attention mechanism is too computationally expensive for the high-dimensional Earth observation data.
How to design a space-time Transformer that is good at predicting the future of the Earth systems is largely an open problem to the community.


In this paper, we propose \emph{Earthformer}, a space-time Transformer for Earth system forecasting. To better explore the design of space-time attention, we propose \emph{Cuboid Attention}, which is a generic building block for efficient space-time attention. The idea is to decompose the input tensor to non-overlapping cuboids and apply cuboid-level self-attention in parallel. Since we limit the $O(N^2)$ self-attention inside the local cuboids, the overall complexity is greatly reduced. Different types of correlations can be captured via different cuboid decompositions. By stacking multiple cuboid attention layers with different hyperparameters, we are able to subsume several previously proposed video Transformers~\cite{ho2019axial, liu2021video, bertasius2021space} as special cases, and also come up with new attention patterns that were not studied before. A limitation of this design is the lack of a mechanism for the local cuboids to communicate with each other. Thus, we introduce a collection of global vectors that attend to all the local cuboids, thereby gathering the overall status of the system. By attending to the global vectors, the local cuboids can grasp the general dynamics of the system and share information with each other.

To verify the effectiveness of cuboid attention and figure out the best design under the Earth system forecasting scenario, we conducted extensive experiments on two synthetic datasets: the MovingMNIST~\cite{shi2015convolutional} dataset and a newly proposed \nbody{} dataset. Digits in the \nbody{} follow the chaotic $3$-body motion pattern~\cite{mj2006three}, which makes the dataset not only more challenging than MovingMNIST but also more relevant to Earth system forecasting. The synthetic experiments reveal the following findings: 1) stacking cuboid attention layers with the Axial attention pattern is both efficient and effective, achieving the best overall performance, 2) adding global vectors provides consistent performance gain without increasing the computational cost, 3) adding hierarchy in the encoder-decoder architecture can improve performance. 
Based on these findings, we figured out the optimal design for Earthformer and made comparisons with other baselines on the SEVIR~\cite{veillette2020sevir} benchmark for precipitation nowcasting and the ICAR-ENSO dataset~\cite{ham2019deep} for ENSO forecasting. Experiments show that Earthformer achieves state-of-the-art (SOTA) performance on both tasks.




\section{Related Work}

\paragraph{Deep learning architectures for Earth system forecasting}
Conventional DL models for Earth system forecasting are based on CNN and RNN. U-Net with either 2D CNN or 3D CNN have been used for precipitation nowcasting~\cite{veillette2020sevir},  Seasonal Arctic Sea ice prediction~\cite{andersson2021seasonal}, and ENSO forecasting~\cite{ham2019deep}. Shi et al.~\cite{shi2015convolutional} proposed the ConvLSTM network that combines CNN and LSTM for precipitation nowcasting. Wang et al.~\cite{wang2022predrnn} proposed PredRNN which adds the spatiotemporal memory flow structure to ConvLSTM. To better learn long-term high-level relations, Wang et al.~\cite{wang2018eidetic} proposed E3D-LSTM that integrates 3D CNN to LSTM. To disentangle PDE dynamics from unknown complementary information, PhyDNet~\cite{guen2020disentangling} incorporates a new recurrent physical cell to perform PDE-constrained prediction in latent space.  Espeholt et al.~\cite{espeholt2021skillful} proposed MetNet-2 that outperforms HREF for forecasting precipitation. The architecture is based on ConvLSTM and dilated CNN. Very recently, there are works that tried to apply Transformer for solving Earth system forecasting problems. Pathak et al.~\cite{pathak2022fourcastnet} proposed the FourCastNet for global weather forecasting, which is based on Adaptive Fourier Neural Operators (AFNO)~\cite{guibas2021adaptive}. Bai et al.~\cite{bai2022rainformer} proposed Rainformer for precipitation nowcasting, which is based on an architecture that combines CNN and Swin-Transformer~\cite{liu2021swin}. In our experiments, we can see that Earthformer outperforms Rainformer.

\paragraph{Space-time Transformers for video modeling}
Inspired by the success of ViT~\cite{dosovitskiy2020image} for image classification, space-time Transformer is adopted for improved video understanding.
In order to bypass the huge memory consumption brought by joint spatiotemporal attention, several pioneering work proposed efficient alternatives, such as divided attention~\cite{bertasius2021space}, axial attention~\cite{ho2019axial,bertasius2021space}, factorized encoder~\cite{neimark2021video,arnab2021vivit} and separable attention~\cite{zhang2021vidtr}. 
Beyond minimal adaptation from ViT, some recent work introduced more prior to the design of space-time transformers, including trajectory~\cite{patrick2021keeping}, multi-scale~\cite{liu2021video,fan2021multiscale} and multi-view~\cite{yan2022multiview}. However, no prior work focuses on exploring the design of space-time Transformers for Earth system forecasting.

\paragraph{Global and local attention in vision Transformers}
To make self-attention more efficient in terms of both memory consumption and speed, recent works have adapted the essence of CNN to perform local attention in transformers~\cite{TNT,ying2021transformers}. HaloNets~\cite{vaswani2021scaling} develops a new self-attention model family consisting of simple local self-attention and convolutional hybrids, which outperform both CNN and vanilla ViT on a range of downstream vision tasks. GLiT~\cite{chen2021glit} introduces a locality module and uses neural architecture search to find an efficient backbone. Focal transformer~\cite{yang2021focal} proposes focal self-attention that can incorporate both fine-grained local and coarse-grained global interactions. 
However, these architectures are not directly applicable to spatiotemporal forecasting. Besides, they are also different from our design because we keep $K$ global vectors to summarize the statistics of the dynamic system and connect the local cuboids; experiments show that such a global vector design is crucial to successful spatiotemporal forecasting. 

\section{Model}
\label{sec:model}
\begin{figure}[!tb]
    \centering
    \begin{minipage}{0.55\textwidth}
     \hspace*{-0.5cm}
        \centering
        \includegraphics[width=1.1\textwidth]{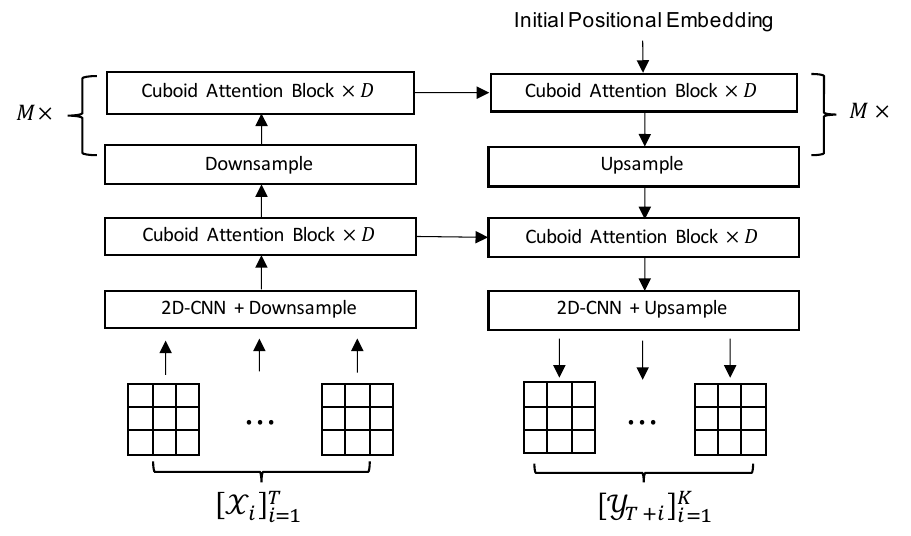}
        \caption{Illustration of the Earthformer architecture. It is a hierarchical Transformer encoder-decoder based on cuboid attention. The input sequence has length $T$ and the target sequence has length $K$. ``$\times D$'' means to stack $D$ cuboid attention blocks with residual connection. ``$M\times$'' means to have $M$ layers of hierarchies.} 
        \label{fig:enc_dec}
    \end{minipage}
     \hspace*{0.5cm}
    \begin{minipage}{0.4\textwidth}
        \centering
        \includegraphics[width=0.95\textwidth]{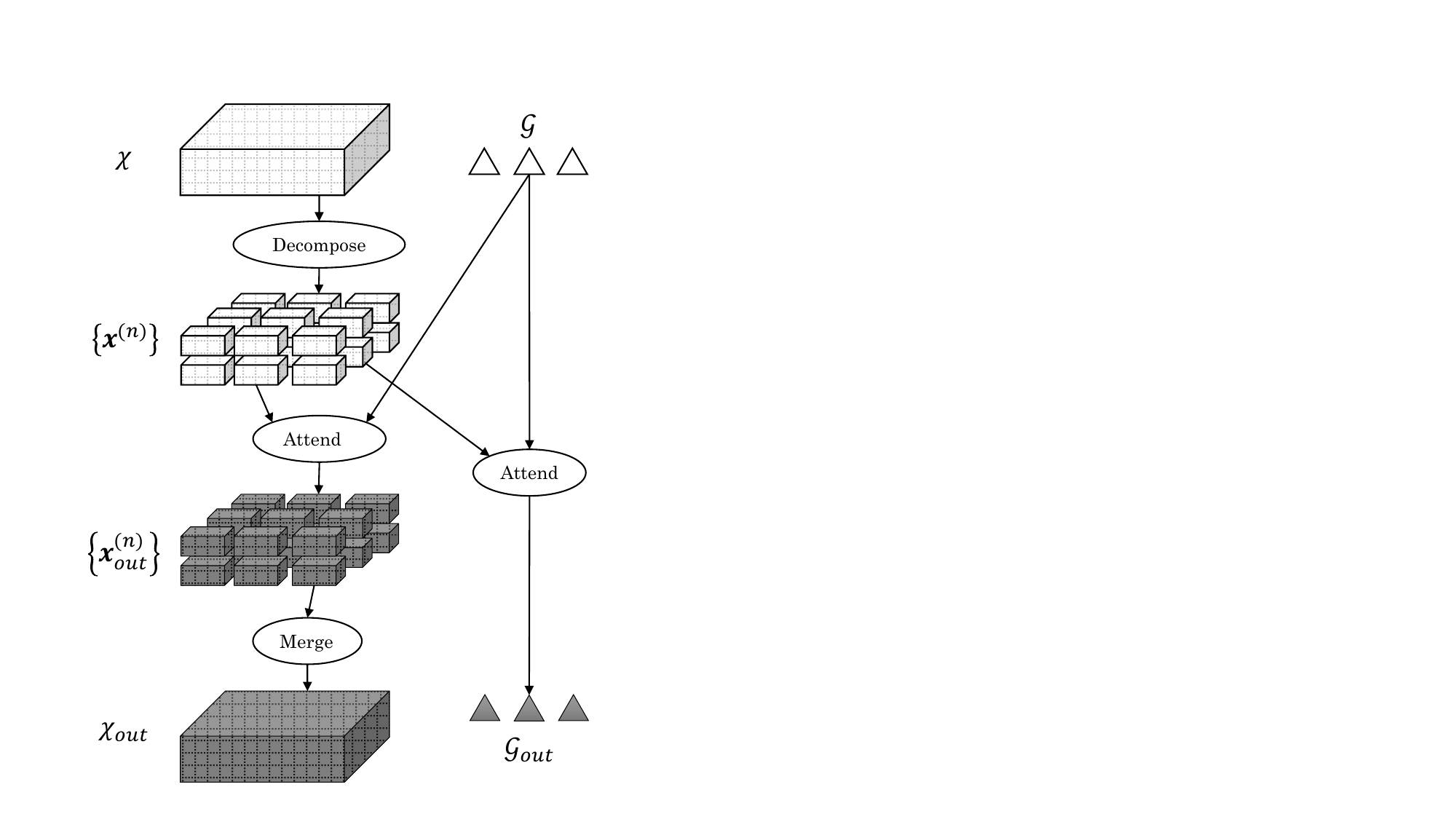}
        \caption{Illustration of the cuboid attention layer with global vectors.}
        \label{fig:cuboid_attention}
    \end{minipage}
    \vspace{-1em}
\end{figure}

Similar to previous works~\cite{shi2015convolutional, veillette2020sevir, bai2022rainformer}, we formulate Earth system forecasting as a spatiotemporal sequence forecasting problem. The Earth observation data, such as radar echo maps from NEXRAD~\cite{heiss1990nexrad} and climate data from CIMP6~\cite{eyring2016overview}, are represented as a spatiotemporal sequence
$[\mathcal{X}_{i}]_{i=1}^T$, $\mathcal{X}_i \in \mathbb{R}^{H\times W\times C_{\text{in}}}$. 
Based on these observations, the model predicts the $K$-step-ahead future
$[\mathcal{Y}_{T+i}]_{i=1}^K$, $\mathcal{Y}_{T+i} \in \mathbb{R}^{H\times W\times C_{\text{out}}}$. 
Here, $H, W$ denote the spatial resolution, and $C_{\text{in}}, C_{\text{out}}$ denote the number of measurements available at each space-time coordinate from the input and target sequences, respectively. As illustrated in~\figref{fig:enc_dec}, our proposed \emph{Earthformer} is a hierarchical Transformer encoder-decoder based on \emph{Cuboid Attention}. The input observations are encoded as a hierarchy of hidden states and then decoded to the prediction target. In what follows, we will present the detailed design of cuboid attention and the hierarchical encoder-decoder architecture adopted in Earthformer.

\subsection{Cuboid Attention}


\begin{figure}
     \centering
     \captionsetup[subfigure]{
        format=hang,
        singlelinecheck=false,
        justification=centering
     }  
     \hspace*{-1.0cm}
     \begin{subfigure}[b]{0.20\textwidth}
         \centering
         \includegraphics[width=1.0\textwidth]{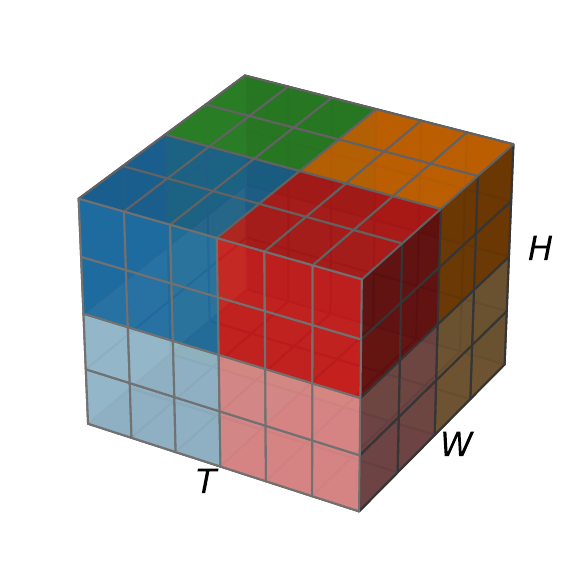}
         \vspace{-0.8cm}
         \caption{$\text{strategy}$=$\mathtt{``local"}$\protect\\ $\text{shift}$=$(0,0,0)$}
     \end{subfigure}
     \qquad
     \begin{subfigure}[b]{0.20\textwidth}
         \centering
         \includegraphics[width=1.0\textwidth]{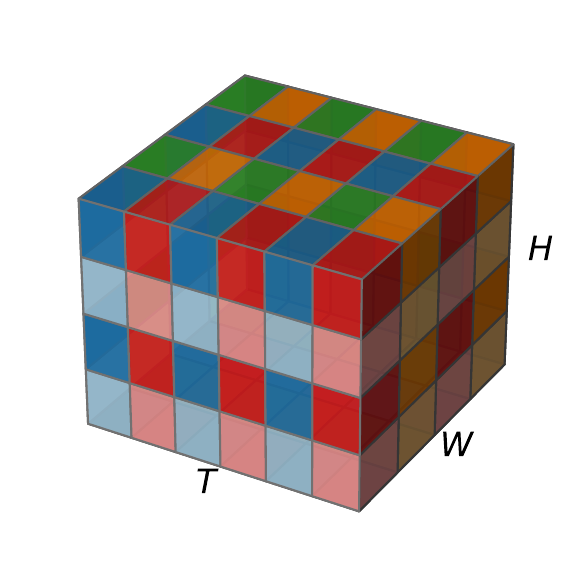}
         \vspace{-0.8cm}
         \caption{$\text{strategy}$=$\mathtt{``dilated"}$\protect\\ $\text{shift}$=$(0,0,0)$}
     \end{subfigure}
     \qquad
     \begin{subfigure}[b]{0.20\textwidth}
         \centering
         \includegraphics[width=1.0\textwidth]{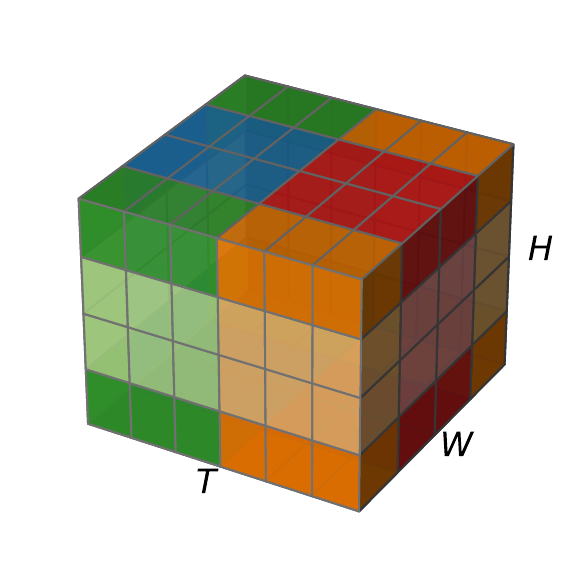}
         \vspace{-0.8cm}
         \caption{$\text{strategy}$=$\mathtt{``local"}$\protect\\ $\text{shift}=(0,1,1)$}
     \end{subfigure}
     \caption{Illustration of cuboid decomposition strategies when the input shape is $(T, H, W) = (6, 4, 4)$, and cuboid size $(b_T, b_H, b_W) = (3, 2, 2)$. Cells that have the same color belong to the same cuboid and will attend to each other. $\text{shift} = (0,1,1)$ shifts the cuboid decomposition by $1$ pixel along height and width dimensions. $\text{strategy}=\mathtt{``local"}$ means to aggregate contiguous $(b_T, b_H, b_W)$ pixels as a cuboid. $\text{strategy}=\mathtt{``dilated"}$ means to aggregate pixels every $\lceil \frac{T}{b_T}\rceil$ ($\lceil \frac{H}{b_H}\rceil$, $\lceil \frac{W}{b_W}\rceil$) steps along time (height, width) dimension. (Best viewed in color).}
     \vspace{-1em}
     \label{fig:cuboid_pattern_illustration}
\end{figure}


Compared with images and text, spatiotemporal data in Earth systems usually have higher dimensionality. As a consequence, applying Transformers to this task is challenging.
For example, for a $3$D tensor with shape $(T, H, W)$, the complexity of the vanilla self-attention is $O(T^2 H^2 W^2)$ and can be computationally infeasible. Previous literature 
proposed various structure-aware space-time attention mechanisms to reduce the complexity~\cite{ho2019axial,liu2021video,bertasius2021space}.

These space-time attention mechanisms share the common design of stacking multiple elementary attention layers that focus on different types of data correlations (e.g., temporal correlation and spatial correlation). Steming from this observation, we propose the generic cuboid attention layer that involves three steps: ``decompose'', ``attend'', and ``merge''.

\paragraph{Decompose}
We first decompose the input spatiotemporal tensor $\mathcal{X}\in\mathbb{R}^{T\times H\times W\times C}$ into a sequence of cuboids $\{\bm{x}^{(n)}\}$.
\begin{align}
    \centering
    \{\bm{x}^{(n)}\} &= \mathtt{Decompose}(\mathcal{X}, \text{cuboid\_size}, \text{strategy}, \text{shift}),
    \label{eq:cuboid_decompose}
\end{align}
where $\text{cuboid\_size}=(b_T,b_H,b_W)$ is the size of the local cuboid, $\text{strategy}\in\{\mathtt{``local"}, \mathtt{``dilated"}\}$ controls whether to adopt the local decomposition strategy or the dilated decomposition strategy~\cite{bertasius2021space}, $\text{shift} = (s_T, s_H, s_W)$ is the window shift offset~\cite{liu2021swin}.
\figref{fig:cuboid_pattern_illustration} provides three examples showing how an input tensor will be decomposed following different hyperparameters of $\mathtt{Decompose}(\cdot)$.
There are a total number of $\lceil \frac{T}{b_T}\rceil\lceil \frac{H}{b_H}\rceil\lceil \frac{W}{b_W}\rceil $ cuboids in $\{\bm{x}^{(n)}\}$. To simplify the notation, we assume that $T, H, W$ are divisible by $b_T, b_H, b_W$. In the implementation, we pad the input tensor if it is not divisible.

Assume $\bm{x}^{(n)}$ is the $(n_T, n_H, n_W)$-th cuboid in $\{\bm{x}^{(n)}\}$. 
The $(i, j, k)$-th element of $\bm{x}^{(n)}$ can be mapped to the $(i^\prime, j^\prime, k^\prime)$-th element of $\mathcal{X}$ via~\eqnref{eq:cuboid_index_map_l} if the strategy is ``local'' or~\eqnref{eq:cuboid_index_map_d} if the strategy is ``dilated''. 

\begin{minipage}{0.49\textwidth}
\begin{align}
    \centering
    \begin{aligned}
    i^\prime&\leftrightarrow s_T+b_T(n_T-1)+i\mod T \\
    j^\prime&\leftrightarrow s_H+b_H(n_H-1)+j\mod H \\
    k^\prime&\leftrightarrow s_W+b_W(n_W-1)+k\mod W
    \end{aligned}
    \label{eq:cuboid_index_map_l}
\end{align}
\end{minipage}
\begin{minipage}{0.49\textwidth}
\begin{align}
    \centering
    \begin{aligned}
    i^\prime&\leftrightarrow s_T+b_T(i-1)+n_T\mod T \\
    j^\prime&\leftrightarrow s_H+b_H(j-1)+n_H\mod H \\
    k^\prime&\leftrightarrow s_W+b_W(k-1)+n_W\mod W
    \end{aligned}
    \label{eq:cuboid_index_map_d}
\end{align}
\end{minipage}

Since the mapping is bijective, one can then map the elements from $\mathcal{X}$ to $\{\bm{x}^{(n)}\}$ via the inverse operation. 

\paragraph{Attend} 
After decomposing the input tensor into a sequence of non-overlapping cuboids $\{\bm{x}^{(n)}\}$, we apply self-attention within each cuboid in parallel.

\begin{align}
    \centering
    \bm{x}^{(n)}_{\text{out}} = \mathtt{Attention}_{\Theta}(\bm{x}^{(n)}, \bm{x}^{(n)}, \bm{x}^{(n)}), 1\leq n\leq N.
    \label{eq:cuboid_local_full_attention}
\end{align}

The query, key, and value matrices $\bm{Q}$, $\bm{K}$, and $\bm{V}$ of $\mathtt{Attention}_\theta(\bm{Q}, \bm{K}, \bm{V}) = \mathtt{Softmax}\left((\bm{W_Q}\bm{Q}) (\bm{W_K}\bm{K})^T / \sqrt{C}\right) (\bm{W_V}\bm{V})$ are all flattened versions of $\bm{x}^{(n)}$, and we unravel the resulting matrix back to a $3$D tensor. $\bm{W_Q}$, $\bm{W_K}$ and $\bm{W_V}$ are linear projection weights and are abbreviated together as $\Theta$.
The self-attention parameter $\Theta$ are shared across all cuboids. 
The computational complexity of the ``attend'' step is $O\left(\lceil \frac{T}{b_T}\rceil\lceil \frac{H}{b_H}\rceil\lceil \frac{W}{b_W}\rceil\left(b_T b_H b_W\right)^2\right) \approx O(T H W \cdot b_T b_H b_W)$, which scales linearly with the cuboid size. Since the cuboid size can be much smaller than the size of the input tensor, the layer is more efficient than full attention.

\paragraph{Merge}
$\mathtt{Merge}(\cdot)$ is the inverse operation of $\mathtt{Decompose}(\cdot)$. The sequence of cuboids obtained after the attention step $\{\bm{x}^{(n)}_{\text{out}}\}$ are merged back to the original input shape to produce the final output of cuboid attention, as shown in~\eqnref{eq:cuboid_merge}. The mapping follows the same bijections in~\eqnref{eq:cuboid_index_map_l} and~\eqnref{eq:cuboid_index_map_d}.
\begin{align}
    \centering
    \mathcal{X}_{\text{out}} = \mathtt{Merge}(\{\bm{x}^{(n)}_{\text{out}}\}_n, \text{cuboid\_size}, \text{strategy}, \text{shift}).
    \label{eq:cuboid_merge}
\end{align}
We combine the ``decompose'', ``attend'' and ``merge'' steps described in~\eqnref{eq:cuboid_decompose},\ref{eq:cuboid_local_full_attention},\ref{eq:cuboid_merge} to construct the generic cuboid attention as in~\eqnref{eq:cuboid_attention_overall}.
\begin{align}
    \centering
    \mathcal{X}_{\text{out}} = \mathtt{CubAttn}_\Theta(\mathcal{X}, \text{cuboid\_size}, \text{strategy}, \text{shift}).
    \label{eq:cuboid_attention_overall}
\end{align}

\paragraph{Explore cuboid attention patterns} 
By stacking multiple cuboid attention layers with different choices of ``$\text{cuboid\_size}$'', ``$\text{strategy}$'' and ``$\text{shift}$'', we are able to efficiently explore existing and potentially more effective space-time attention. In this paper, we explore the cuboid attention patterns as listed in~\tabref{table:cuboid_attention_patterns}. From the table, we can see that cuboid attention  subsumes previously proposed space-time attention methods like axial attention, video swin-Transformer, and divided space-time attention. Also, we manually picked the patterns that are reasonable and not computationally expensive as our search space. The flexibility of cuboid attention allows us to conduct Neural Architecture Search (NAS) to automatically search for a pattern but we will leave it as future work.

\begin{table}[!tb]
\vskip -0.4cm
\caption{Configurations of the cuboid attention patterns explored in the paper. 
The input tensor has shape $(T, H, W)$. 
If ``$\text{shift}$'' or ``$\text{strategy}$'' is not given, we use $\text{shift}=\mathtt{(0, 0, 0)}$ and $\text{strategy}=\mathtt{``local"}$ by default.
When stacking multiple cuboid attention layers, each layer will be coupled with layer normalization layers and feed-forward network as in the Pre-LN Transformer~\cite{xiong2020layer}. 
The first row shows the configuration of the generic cuboid attention.} 
\label{table:cuboid_attention_patterns}
\begin{center}
\resizebox{0.88\textwidth}{!}{
\begin{tabular}{ l|c|c }
\toprule[1.5pt]
Name & Configurations & Values\\
\midrule
\midrule
\multirow{3}{*}{Generic Cuboid Attention} & cuboid\_size & $\mathtt{(T_1, H_1, W_1)\ \rightarrow\ (T_2, H_2, W_2)\ \ \rightarrow \dots\rightarrow\ (T_L, H_L, W_L)}$ \\ 
& shift & $\mathtt{(P_1, M_1, M_1)\ \rightarrow\ (P_2, M_2, M_2)\ \ \rightarrow \dots\rightarrow\ (P_L, M_L, M_L)}$ \\
& strategy & $\mathtt{``loc./dil."\rightarrow``loc./dil."\rightarrow\dots\rightarrow``loc./dil."}$ \\
\midrule
\midrule
Axial & cuboid\_size & $\mathtt{(T, 1, 1)\rightarrow (1, H, 1)\rightarrow (1, 1, W)}$ \\ 
\midrule
Divided Space-Time & cuboid\_size & $\mathtt{(T, 1, 1)\rightarrow (1, H, W)}$ \\
\midrule
\multirow{2}{*}{Video-Swin $P\times M$} & cuboid\_size & $\mathtt{(P, M, M)\rightarrow (\ \ P\ \ , \ \ M\ \ , \ \ M\ \ )}$ \\ 
& shift & $\mathtt{(0, 0, 0)\rightarrow(P/2, M/2, M/2)}$ \\
\midrule
\multirow{2}{*}{Spatial Local-Dilate-$M$} & cuboid\_size & $\mathtt{(T, 1, 1) \rightarrow (1, M, M)\ \rightarrow (1, M, M)}$ \\ 
& strategy & $\mathtt{``local" \rightarrow``local"\rightarrow``dilated"}$ \\
\midrule
\multirow{2}{*}{Axial Space Dilate-$M$} & cuboid\_size & $\mathtt{(T, 1, 1)\rightarrow (1, H / M, 1)\rightarrow (1, H / M, 1)\rightarrow (1, 1, W / M)\rightarrow (1, 1, W / M)}$ \\ 
& strategy & $\mathtt{``local"\rightarrow``dilated"\rightarrow``local"\rightarrow``dilated"\rightarrow``local"}$ \\
\bottomrule[1.5pt]
\end{tabular}
} 
\end{center}
\vskip -0.5cm
\end{table}

\subsection{Global Vectors}
One limitation of the previous formulation is that the cuboids do not communicate with each other. This is undesirable because each cuboid is not capable of understanding the global dynamics of the system. Thus, inspired by the \texttt{[CLS]} token adopted in BERT~\cite{devlin2018bert,zaheer2020big}, we propose to introduce a collection of $P$ global vectors $\mathcal{G}\in\mathbb{R}^{P\times C}$ to help cuboids scatter and gather crucial global information. When each cuboid is performing the self-attention, the elements will not only attend to the other elements within the same cuboid but also attend to the global vectors $\mathcal{G}$. 
We revise~\eqnref{eq:cuboid_local_full_attention} to~\eqnref{eq:cuboid_l2g_attention} to enable local-global information exchange. 
We also use~\eqnref{eq:cuboid_g2l_attention} to update the global vectors $\mathcal{G}$ by aggregating the information from all elements of the input tensor $\mathcal{X}$.
%
%
\begin{align}
    \centering
    \bm{x}^{(n)}_{\text{out}} &= \mathtt{Attention}_{\Theta}\left(\bm{x}^{(n)}, \mathtt{Cat}(\bm{x}^{(n)},\mathcal{G}), \mathtt{Cat}(\bm{x}^{(n)},\mathcal{G}) \right), 1\leq n\leq N.
    \label{eq:cuboid_l2g_attention} \\
    \mathcal{G}_{\text{out}} &= \mathtt{Attention}_{\Phi}\left(\mathcal{G}, \mathtt{Cat}(\mathcal{G},\mathcal{X}), \mathtt{Cat}(\mathcal{G}, \mathcal{X})\right).
    \label{eq:cuboid_g2l_attention}
\end{align}

Here, $\mathtt{Cat}(\cdot)$ flattens and concatenates its input tensors. 
By combining~\eqnref{eq:cuboid_decompose},\ref{eq:cuboid_l2g_attention},\ref{eq:cuboid_g2l_attention},\ref{eq:cuboid_merge}, we abbreviate the overall computation of the cuboid attention layer with global vectors as in~\eqnref{eq:cuboid_attention_global_overall}.
\begin{align}
    \centering
    \begin{aligned}
    \mathcal{X}_{\text{out}} &= \mathtt{CubAttn}_{\Theta}(\mathcal{X}, \mathcal{G}, \text{cuboid\_size}, \text{strategy}, \text{shift}), \\
    \mathcal{G}_{\text{out}} &= \mathtt{Attn}_{\Phi}^{\text{global}}(\mathcal{G}, \mathcal{X}).
    \end{aligned}
    \label{eq:cuboid_attention_global_overall}
\end{align}
The additional complexity caused by the global vectors is approximately $O\left(THW\cdot P + P^2\right)$. Given that $P$ is usually small (in our experiments, $P$ is at most $8$), the computational overhead induced by the global structure is negligible. The architecture of the cuboid attention layer is illustrated in~\figref{fig:cuboid_attention}.


\subsection{Hierarchical Encoder-Decoder Architecture}
\label{sec:hierarchical_enc_dec}

Earthformer adopts a hierarchical encoder-decoder architecture illustrated in~\figref{fig:enc_dec}. The hierarchical architecture gradually encodes the input sequence to multiple levels of representations and generates the prediction via a coarse-to-fine procedure. Each hierarchy stacks $D$ cuboid attention blocks. The cuboid attention block in the encoder uses one of the patterns described in~\tabref{table:cuboid_attention_patterns}, and each cuboid block in the decoder adopts the ``Axial'' pattern. To reduce the spatial resolution of the input to cuboid attention layers, we include a pair of initial downsampling and upsampling modules that consist of stacked 2D-CNN and Nearest Neighbor Interpolation (NNI) layers.
Different from other papers that adopt Transformer for video prediction~\cite{ho2019axial, rakhimov2020latent}, we generate the predictions in a non-auto-regressive fashion rather than an auto-regressive patch-by-patch fashion. This means that our decoder directly generates the predictions from the initial learned positional embeddings. We also conducted experiments with an auto-regressive decoder based on visual codebook~\cite{razavi2019generating}. However, the auto-regressive decoder underperforms the non-auto-regressive decoder in terms of forecasting skill scores. The comparison between non-auto-regressive decoder and auto-regressive decoder is shown in Appendix~\ref{sec:nar-vs-ar}.

\section{Experiments}
We first conducted experiments on two synthetic datasets, MovingMNIST and a newly proposed \nbody{}, to verify the effectiveness of Earthformer and conduct ablation study on our design choices. Results on these two datasets lead to the following findings: 1) Among all patterns listed in~\tabref{table:cuboid_attention_patterns}, ``Axial'' achieves the best overall performance; 2) Global vectors bring consistent performance gain with negligible increase in computational cost; 3) Using a hierarchical coarse-to-fine structure can boost the performance.
Based on these findings, we figured out the optimal design of Earthformer and compared it with other state-of-the-art models
on two real-world datasets: SEVIR~\cite{veillette2020sevir} and ICAR-ENSO\footnote{Dataset available at \url{https://tianchi.aliyun.com/dataset/dataDetail?dataId=98942}}. On both datasets, Earthformer achieved the best overall performance. The statistics of all the datasets used in the experiments are shown in~\tabref{table:datasets_info}.
We normalized the data to the range $[0, 1]$ and trained all the models with the Mean-Squared Error (MSE) loss. More implementation details are shown in Appendix~\ref{sec:implementatoin-details}.


\begin{table}[!tb]
\vskip -0.4cm
\caption{Statistics of the datasets used in the experiments.}
\label{table:datasets_info}
	\begin{center}
	\resizebox{0.60\textwidth}{!}{
	\begin{tabular}{l|ccc|cc|c}
	\toprule[1.5pt]
	\multirow{2}{*}{Dataset}    & \multicolumn{3}{|c|}{Size}        & \multicolumn{2}{|c|}{Seq. Len.}   & Spatial Resolution \\
	\addlinespace[0.03cm]
	\cline{2-7}
	\addlinespace[0.05cm]
	                            & train     & val       & test      & in            & out               & $H\times W$\\ 
	\midrule\midrule
	MovingMNIST                 &\ \ 8,100  &\ \ 900    &\ \ 1,000  & 10            & 10                & $64\times64$ \\
	\nbody{}                    & 20,000    & 1,000     &\ \ 1,000  & 10            & 10                & $64\times64$ \\
	SEVIR                       & 35,718    & 9,060     &   12,159  & 13            & 12                & $384\times384$ \\
	ICAR-ENSO                   &\ \ 5,205  &\ \ 334    &\ \ 1,667  & 12            & 14                & $24\times48$ \\
	\bottomrule[1.5pt]
	\end{tabular}
	}
	\end{center}
	\vskip -0.5cm
\end{table}

\subsection{Experiments on Synthetic Datasets}
\label{sec:exp_synthetic}

\paragraph{MovingMNIST}
We follow~\cite{srivastava2015unsupervised} to use the public MovingMNIST dataset\footnote{MovingMNIST: \url{https://github.com/mansimov/unsupervised-videos}}. 
The dataset contains 10,000 sequences. Each sequence shows 2 digits moving inside a $64\times 64$ frame. We split the dataset to use 8,100 samples for training, 900 samples for validation and 1,000 samples for testing. The task is to predict the future 10 frames for each sequence conditioned on the first 10 frames.

\paragraph{\nbody{}}
The Earth is a complex system in which an extremely large number of variables interact with each other.
Compared with the Earth system, the dynamics of the synthetic MovingMNIST dataset, in which the digits move independently with constant speed, is over-simplified.
Thus, achieving good performance on MovingMNIST does not imply that the model is capable of modeling complex interactions in the Earth system.
On the other hand, the real-world Earth observation data, though experiencing rapid development, are still noisy and may not provide useful insights for model development.
Therefore, we extend MovingMNIST to \nbody{}, where $N$ digits are moving with the $N$-body motion pattern inside a $64 \times 64$ frame. Each digit has its mass and is subjected to the gravity from other digits.
We choose $N=3$ in the experiments so that the digits will follow the chaotic $3$-body motion~\cite{mj2006three}. The highly non-linear interactions in \nbody{} make it much more challenging than the original MovingMNIST. We generate 20,000 sequences for training, 1,000 for validation and 1,000 for testing. Perceptual examples of the dataset can be found at the first two rows of~\figref{fig:qualitative_nbody}. In Appendix~\ref{sec:chaos-nbody}, we demonstrate the chaotic behavior of \nbody{}.

\paragraph{Hierarchical v.s.\ non-hierarchical}
We choose ``Axial'' without global vectors as our cuboid attention pattern and compare the performance of non-hierarchical and hierarchical architectures on MovingMNIST. 
The ablation study on the importance of adopting a hierarchical encoder-decoder is shown in~\tabref{table:moving_mnist_hierarchy_scores}. We can see that the hierarchical architecture has similar FLOPS with the non-hierarchical architectures while being better in MSE. This observation is consistent as we increase the depth until the performance saturates.

\begin{table}[!tb]
\vskip -0.4cm
\caption{Ablation study on the importance of adopting a hierarchical encoder-decoder. We conducted experiments on MovingMNIST. ``Depth $D$'' means the model stacks $D$ cuboid attention blocks and there is no hierarchical structure. ``Depth $D1,D2$'' means the model stacks $D1$ cuboid attention blocks, applies the pooling layer, and stacks another $D2$ cuboid attention blocks. }
\label{table:moving_mnist_hierarchy_scores}
	\begin{center}
	\resizebox{0.55\textwidth}{!}{
	\begin{tabular}{l|c|c|ccc}
	\toprule[1.5pt]
	\multirow{2}{*}{Model}  & \multirow{2}{*}{\#Param. (M)} & \multirow{2}{*}{GFLOPS}& \multicolumn{3}{|c}{Metrics}   \\
 						    &				                &                        & MSE  $\downarrow$    & MAE  $\downarrow$ & SSIM $\uparrow$ \\
	\midrule\midrule
	Depth 2				& 1.4     & 17.9	& 63.80				& 140.6				& 0.8324				\\
	Depth 4				& 3.1     & 36.3	& 52.46				& 114.8				& 0.8685				\\
	Depth 6				& 4.9     & 54.6	& 50.49				& 110.0				& 0.8738				\\
	Depth 8				& 6.6     & 73.0	& 48.04				& 104.6				& 0.8797				\\
	\midrule
	Depth 1,1			& 1.4     & 11.5	& 60.99				& 135.7				& 0.8388				\\
    Depth 2,2			& 3.1     & 18.9    & 50.41				& 106.9				& 0.8805				\\
	Depth 3,3			& 4.9     & 26.3	& \underline{47.69}	& \textbf{100.1}	& \textbf{0.8873}	    \\
	Depth 4,4			& 6.6     & 33.7    & \textbf{46.91}	& \underline{101.5}	& \underline{0.8825}	\\
	\bottomrule[1.5pt]
	\end{tabular}
	}
	\end{center}
	\vskip -0.5cm
\end{table}

\paragraph{Cuboid pattern search}
The design of cuboid attention greatly facilitates the search for optimal space-time attention. 
We compare the patterns listed in~\tabref{table:cuboid_attention_patterns} on both MovingMNIST and \nbody{} to investigate the effectiveness and efficiency of different space-time attention methods on spatiotemporal forecasting tasks. 
Besides the previously proposed space-time attention methods, we also include new configurations that are reasonable and not computationally expensive in our search space.   For each pattern, we also compare the variant that uses global vectors. Results are summarized in~\tabref{table:depth44_pattern_search}. We find that the ``Axial'' pattern is both effective and efficient and adding global vectors improves performance for all patterns while having similar FLOPS. We thus pick ``Axial + global'' as the pattern in Earthformer when conducting experiments on real-world datasets.

\begin{table}[!tb]
\caption{Ablation study of different cuboid attention patterns and the effect of global vectors on MovingMNIST and \nbody{}. The variant that achieved the best performance is in boldface while the second best is underscored. We also compared the performance of the cuboid attention patterns with and without global vectors and highlight the better one with grey background.}
\label{table:depth44_pattern_search}
	\begin{center}
	\resizebox{0.83\textwidth}{!}{
	\begin{tabular}{l|c|c|ccc|ccc}
	\toprule[1.5pt]
	\multirow{2}{*}{Model}  & \multirow{2}{*}{\#Param. (M)}& \multirow{2}{*}{GFLOPS}& \multicolumn{3}{|c}{MovingMNIST}   & \multicolumn{3}{|c}{\nbody{}}  \\
 						    &				&           & MSE  $\downarrow$	    & MAE  $\downarrow$     & SSIM $\uparrow$           & MSE $\downarrow$	& MAE  $\downarrow$ & SSIM $\uparrow$\\
	\midrule\midrule
	Axial				    &\ \ 6.61	    & 33.7		& 46.91				    & 101.5				    & 0.8825			        & 15.89				& \underline{41.38}	& \underline{0.9510}	\\
	+ global $\bigstar$     &\ \ 7.61		& 34.0		& \textbf{\bgray{41.79}}& \textbf{\bgray{92.78}}& \textbf{\bgray{0.8961}}	& \bgray{\textbf{14.82}}& \bgray{\textbf{39.93}}& \bgray{\textbf{0.9538}}\\
	\midrule
	DST					    &\ \ 5.70	    & 35.2		& 57.43				    & 118.6				    & 0.8623			        & 18.24				& 45.88				& \bgray{0.9435}		\\
	+ global			    &\ \ 6.37       & 35.5		& \bgray{52.92}	        & \bgray{108.3}         & \bgray{0.8760}	        & \bgray{17.77}		& \bgray{45.84}		& 0.9433				\\
	\midrule
	Video Swin 2x8		    &\ \ 5.66       & 31.1		& 54.45				    & 111.7				    & 0.8715		            & 19.89				& 49.02				& 0.9374				\\
	+ global                &\ \ 6.33       & 31.4		& \bgray{52.70}	        & \bgray{108.5}	        & \bgray{0.8766}            & \bgray{19.53}		& \bgray{48.43}		& \bgray{0.9389}		\\
	\midrule
	Video Swin 10x8		    &\ \ 5.89       & 39.2		& 63.34				    & 125.3				    & 0.8525			        & 23.35				& 53.17				& 0.9274				\\
	+ global			    &\ \ 6.56       & 39.4		& \bgray{62.15}	        & \bgray{123.4}	        & \bgray{0.8541}	        & \bgray{22.81}		& \bgray{52.94}		& \bgray{0.9293}		\\
	\midrule
	Spatial Local-Global 2  &\ \ 6.61       & 33.3     & 59.88				    & \bgray{122.1}         & \bgray{0.8572}	        & 23.24				& 54.63				& 0.9263				\\
	+ global                &\ \ 7.61	    & 33.7	    & \bgray{59.42}         & 122.9				    & 0.8565			        & \bgray{21.88}		& \bgray{52.49}		& \bgray{0.9305}		\\
	\midrule
	Spatial Local-Global 4  &\ \ 6.61	    & 33.5	    & 58.72				    & 118.5				    & \bgray{0.8600}	        & 21.02				& 49.82				& 0.9344				\\
	+ global                &\ \ 7.61	    & 33.9	    & \bgray{54.84}         & \bgray{115.5}         & 0.8585			        & \bgray{19.82}		& \bgray{48.12}		& \bgray{0.9371}		\\
	\midrule
	Axial Space Dilate 2    &\ \ 8.59	    & 41.8	    & 50.11				    & 104.4				    & 0.8814			        & 15.97				& 42.19				& 0.9494				\\
	+ global                & 10.30         & 42.4   	& \bgray{46.86}	        & \bgray{98.95}	        & \bgray{0.8884}	        & \bgray{\underline{15.73}}& \bgray{41.85}& \bgray{0.9510}		\\
	\midrule
	Axial Space Dilate 4    &\ \ 8.59	    & 41.6		& 47.40				    & 99.31				    & 0.8865			        & 19.49				& 51.04				& 0.9352				\\
	+ global                & 10.30	        & 42.2	    & \bgray{\underline{45.11}}& \bgray{\underline{95.98}}& \bgray{\underline{0.8928}}& \bgray{17.91}& \bgray{46.35}	& \bgray{0.9440}				\\
	\bottomrule[1.5pt]
	\end{tabular}
	} 
	\end{center}
	\vspace{-1.5em}
\end{table}

\paragraph{Comparison to the state of the art}
We evaluate six spatiotemporal forecasting algorithms: UNet~\cite{veillette2020sevir}, ConvLSTM~\cite{shi2015convolutional}, PredRNN~\cite{wang2022predrnn}, PhyDNet~\cite{guen2020disentangling}, E3D-LSTM~\cite{wang2018eidetic} and Rainformer~\cite{bai2022rainformer}. The results are in~\tabref{table:synthetic_scores}. Note that the MovingMNIST performance on several papers~\cite{guen2020disentangling} is obtained by training the model with on-the-fly generated digits while we pre-generate the digits and train all models on a fixed dataset. Comparing the numbers in the table with numbers shown in these papers are not fair. We train all baselines from scratch on both MovingMNIST and \nbody{} using the default hyperparameters and configurations in their officially released code\footnote{Except for Rainformer which originally has 212M parameters and thus suffers from overfitting severely.}.

\begin{table}[!tb]
\setlength{\tabcolsep}{3.5pt}
\caption{Comparison of Earthformer with baselines on MovingMNIST and \nbody{}.}
\label{table:synthetic_scores}
	\begin{center}
	\resizebox{0.8\textwidth}{!}{
	\begin{tabular}{l|c|c|ccc|ccc}
	\toprule[1.5pt]
    \multirow{2}{*}{Model}& \multirow{2}{*}{\#Param. (M)}& \multirow{2}{*}{GFLOPS}  & \multicolumn{3}{|c}{MovingMNIST}        & \multicolumn{3}{|c}{\nbody{}}  \\
 						                &           &		    & MSE $\downarrow$  & MAE $\downarrow$  & SSIM $\uparrow$   & MSE $\downarrow$  & MAE $\downarrow$  & SSIM $\uparrow$	\\
	\midrule\midrule
	UNet~\cite{veillette2020sevir}      & 16.6		&\quad0.9   & 110.4				& 249.4				& 0.6170		    & 38.90				& 94.29				& 0.8260\\
	ConvLSTM~\cite{shi2015convolutional}& 14.0	    &\ \ 30.1   & 62.04				& 126.9				& 0.8477            & 32.15				& 72.64				& 0.8886\\
	PredRNN~\cite{wang2022predrnn}      & 23.8		& 232.0     & 52.07				& 108.9				& \underline{0.8831}& 21.76				& 54.32				& 0.9288\\
	PhyDNet~\cite{guen2020disentangling}&\ \ 3.1	&\ \ 15.3	& 58.70				& 124.1				& 0.8350            & 28.97				& 78.66				& 0.8206\\
	E3D-LSTM~\cite{wang2018eidetic}     & 12.9		& 302.0	    & 55.31				& 101.6				& 0.8821			& 22.98				& 62.52				& 0.9131\\
	Rainformer~\cite{bai2022rainformer} & 19.2		&\quad 1.2  & 85.83				& 189.2				& 0.7301            & 38.89				& 96.47				& 0.8036\\
	\midrule
	Earthformer w/o global	            &\ \ 6.6    &\ \ 33.7	& \underline{46.91}	& \underline{101.5}	& 0.8825            & \underline{15.89}	& \underline{41.38}	& \underline{0.9510} \\
	Earthformer			                &\ \ 7.6	&\ \ 34.0	& \textbf{41.79}	& \textbf{92.78}	& \textbf{0.8961}	& \textbf{14.82}	& \textbf{39.93}	& \textbf{0.9538}\\
	\bottomrule[1.5pt]
	\end{tabular}
	}  
	\end{center}
	\vspace{-1.5em}
\end{table}

\begin{figure}[!tb]
    \centering
    \vskip -0.4cm
    \includegraphics[width=0.8\textwidth]{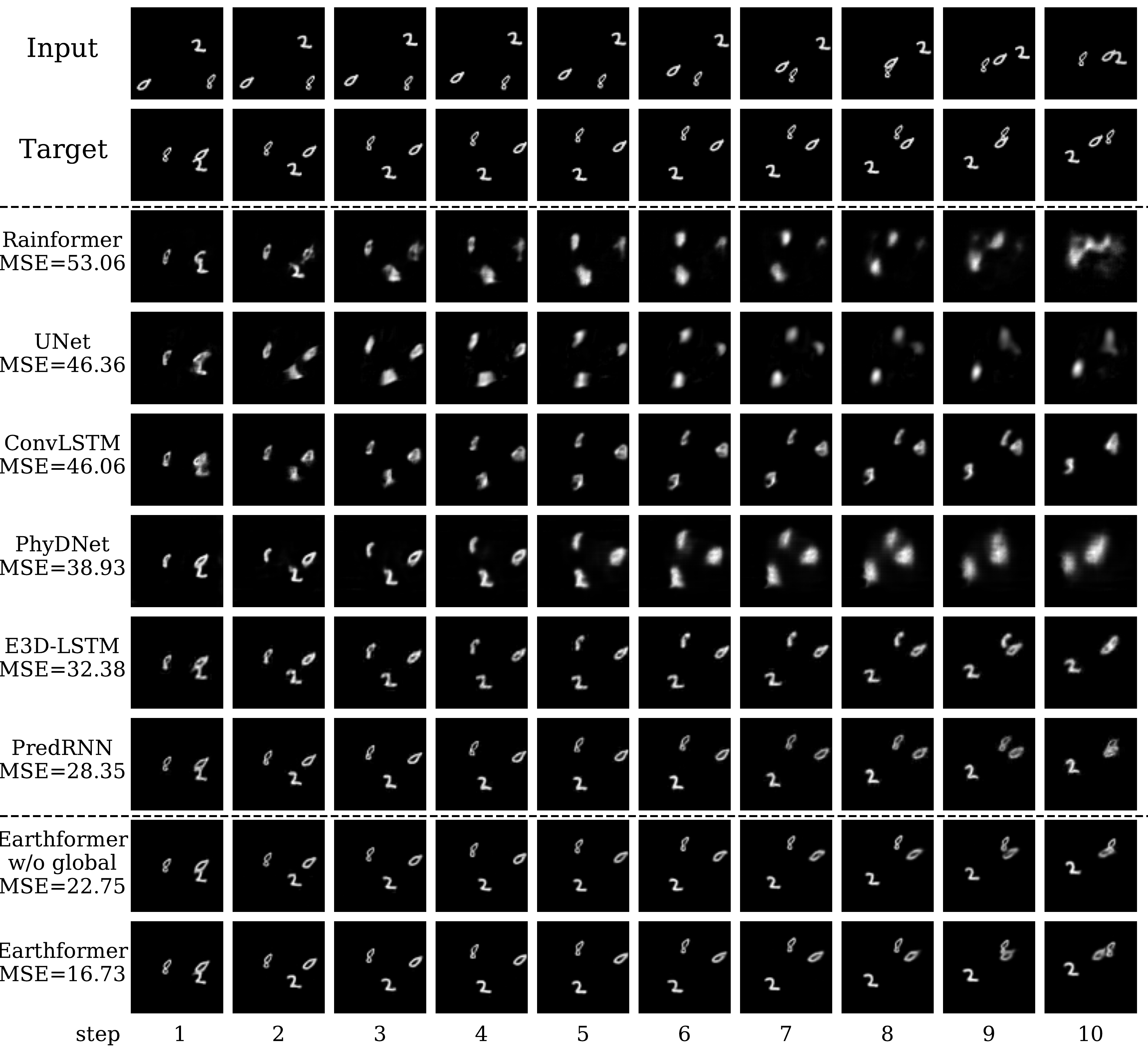}
    \vskip -0.2cm
    \caption{A set of examples showing the perceptual quality of the predictions on the \nbody{} test set. 
    From top to bottom: input frames, target frames, predictions by Rainformer~\cite{bai2022rainformer}, UNet~\cite{veillette2020sevir}, ConvLSTM~\cite{shi2015convolutional}, PhyDNet~\cite{guen2020disentangling}, E3D-LSTM~\cite{wang2018eidetic}, PredRNN~\cite{wang2022predrnn}, Earthformer without using global vectors, Earthformer. The results are sorted according to the MSE.}
    \label{fig:qualitative_nbody}
    \vskip -0.4cm
\end{figure}

\paragraph{Qualitative results on \nbody{}}
\figref{fig:qualitative_nbody} shows the generation results of different methods on a sample sequence from the \nbody{} test set.
The qualitative example demonstrates that our Earthformer is capable of learning long-range interactions among digits and correctly predicting their future motion trajectories. Also, we can see that Earthformer is able to more accurately predict the position of the digits with the help of global vectors.
On the contrary, none of the baseline algorithms that achieved solid performance on MovingMNIST gives the correct and precise position of the digit ``0'' in the last frame.
They either predict incorrect motion trajectories (PredRNN and E3D-LSTM), or generate highly blurry predictions (Rainformer, UNet and PhyDNet) to accommodate the uncertainty about the future.

\subsection{SEVIR Precipitation Nowcasting}
\label{sec:exp_sevir}
Storm EVent ImageRy (SEVIR) \cite{veillette2020sevir} is a spatiotemporally aligned dataset containing over 10,000 weather events. Each event consists of $384\ \mathtt{km}\times384\ \mathtt{km}$ image sequences spanning over 4 hours. Images in SEVIR were sampled and aligned across five different data types: three channels (C02, C09, C13) from the GOES-16 advanced baseline imager, NEXRAD Vertically Integrated Liquid (VIL) mosaics, and GOES-16 Geostationary Lightning Mapper (GLM) flashes. 
The SEVIR benchmark supports scientific research on multiple meteorological applications including precipitation nowcasting, synthetic radar generation, front detection, etc. We adopt SEVIR for benchmarking precipitation nowcasting, i.e., to predict the future VIL up to 60 minutes (12 frames) given 65 minutes context VIL (13 frames). ~\figref{fig:sevir_example} shows an example of VIL observation sequences in SEVIR.

Besides MSE, we also include the $\mathtt{Critical\ Success\ Index}$ ($\mathtt{CSI}$), which is commonly used in precipitation nowcasting and is defined as $\mathtt{CSI} = \frac{\mathtt{\#Hits}}{\mathtt{\#Hits}+\mathtt{\#Misses}+\mathtt{\#F.Alarms}}$. To count the $\mathtt{\#Hits}$ (truth=1, pred=1), $\mathtt{\#Misses}$ (truth=1, pred=0) and $\mathtt{\#F.Alarms}$ (truth=0, pred=1), 
the prediction and the ground-truth are rescaled back to the range 0-255 and binarized at thresholds $[16, 74, 133, 160, 181, 219]$. 
We report $\mathtt{CSI}$ at different thresholds and also their mean $\mathtt{CSI}\mbox{-}\mathtt{M}$.

SEVIR is much larger than MovingMNIST and \nbody{} and has higher resolution. We thus slightly adjust the configurations of baselines based on those for MovingMNIST for fair comparison. Detailed configurations are shown in Appendix~\ref{sec:implementatoin-details}. 
The experiment results are listed in~\tabref{table:sevir_csi}. 
Earthformer consistently outperforms baselines on almost all metrics and brings significant performance gain especially at high thresholds like $\mathtt{CSI}\mbox{-}219$, which are more valued by the communities.

\begin{table}[!tb]
\vskip -0.4cm
\caption{Performance comparison on SEVIR. We include $\mathtt{Critical\ Success\ Index}$ ($\mathtt{CSI}$) besides MSE as evaluation metrics. The $\mathtt{CSI}$, a.k.a intersection over union (IOU), is calculated at different precipitation thresholds and denoted as $\mathtt{CSI}\mbox{-}thresh$.}
\label{table:sevir_csi}
	\begin{center}
	\resizebox{0.90\textwidth}{!}{
	\begin{tabular}{l|c|c|cccccccc}
	\toprule[1.5pt]
	\multirow{2}{*}{Model}& \multirow{2}{*}{\#Param. (M)}& \multirow{2}{*}{GFLOPS} & \multicolumn{8}{|c}{Metrics}\\
 						                &           &       & $\mathtt{CSI}\mbox{-}\mathtt{M}$ $\uparrow$ & $\mathtt{CSI}\mbox{-}219$ $\uparrow$ & $\mathtt{CSI}\mbox{-}181$ $\uparrow$ & $\mathtt{CSI}\mbox{-}160$ $\uparrow$ & $\mathtt{CSI}\mbox{-}133$ $\uparrow$ & $\mathtt{CSI}\mbox{-}74$ $\uparrow$ & $\mathtt{CSI}\mbox{-}16$ $\uparrow$ & MSE ($10^{-3}$) $\downarrow$\\
	\midrule\midrule
	Persistence			                & -         & -     & 0.2613				& 0.0526				& 0.0969				& 0.1278				& 0.2155				& 0.4705				& 0.6047				& 11.5338				\\
	UNet~\cite{veillette2020sevir}      &\ \ 16.6   &\ \ 33	& 0.3593				& 0.0577				& 0.1580				& 0.2157				& 0.3274				& 0.6531				& 0.7441				& 4.1119				\\
	ConvLSTM~\cite{shi2015convolutional}&\ \ 14.0   & 527	& 0.4185				& 0.1288				& 0.2482				& 0.2928				& 0.4052				& 0.6793				& 0.7569				& 3.7532				\\
	PredRNN~\cite{wang2022predrnn}      &\ \ 46.6   & 328	& 0.4080				& 0.1312				& 0.2324				& 0.2767				& 0.3858				& 0.6713				& 0.7507				& 3.9014				\\
	PhyDNet~\cite{guen2020disentangling}&\ \ 13.7   & 701	& 0.3940				& 0.1288				& 0.2309				& 0.2708				& 0.3720				& 0.6556				& 0.7059				& 4.8165				\\
	E3D-LSTM~\cite{wang2018eidetic}     &\ \ 35.6   & 523	& 0.4038				& 0.1239				& 0.2270				& 0.2675				& 0.3825				& 0.6645				& \underline{0.7573}	& 4.1702				\\
	Rainformer~\cite{bai2022rainformer} & 184.0     & 170   & 0.3661				& 0.0831				& 0.1670				& 0.2167				& 0.3438				& 0.6585				& 0.7277				& 4.0272				\\
	\midrule
	Earthformer w/o global	            &\ \ 13.1   & 257	& \underline{0.4356}	& \underline{0.1572}	& \underline{0.2716}	& \underline{0.3138}	& \underline{0.4214}	& \underline{0.6859}	& \textbf{0.7637}		& \underline{3.7002}	\\
	Earthformer			                &\ \ 15.1   & 257	& \textbf{0.4419}		& \textbf{0.1791}		& \textbf{0.2848}		& \textbf{0.3232}		& \textbf{0.4271}		& \textbf{0.6860}		& 0.7513				& \textbf{3.6957}		\\
	\bottomrule[1.5pt]
	\end{tabular}
	}
	\vspace{-1em}
	\end{center}
\end{table}

\subsection{ICAR-ENSO Sea Surface Temperature Anomalies Forecasting}
El Niño/Southern Oscillation (ENSO) has a wide range of associations with regional climate extremes and ecosystem impacts.
ENSO sea surface temperature (SST) anomalies forecasting for lead times up to one year (12 steps) is a valuable and challenging problem. Nino3.4 index, which is the area-averaged SST anomalies across a certain area ($170^\circ$-$120^\circ\mathtt{W}$, $5^\circ\mathtt{S}$-$5^\circ\mathtt{N}$) of the Pacific, serves as a crucial indicator of this climate event. 
The forecast quality is evaluated by the correlation skill~\cite{ham2019deep} of the three-month-moving-averaged Nino3.4 index $\Cnino{} = \frac{\sum_N(\bm{X}-\bar{\bm{X}})(\bm{Y}-\bar{\bm{Y}})}{\sqrt{\sum_N(\bm{X}-\bar{\bm{X}})^2\sum_N(\bm{Y}-\bar{\bm{Y}})^2}}\in\mathbb{R}^K$ calculated on the whole test set of size $N$, where $\bm{Y}\in\mathbb{R}^{N\times K}$ is the ground-truth of $K$-step Nino3.4 index, $\bm{X}\in\mathbb{R}^{N\times K}$ is the corresponding prediction of Nino3.4 index.

ICAR-ENSO consists of historical climate observation and stimulation data provided by Institute for Climate and Application Research (ICAR). We forecast the SST anomalies up to 14 steps (2 steps more than one year for calculating three-month-moving-average) given a context of 12 steps of SST anomalies observations. 
~\tabref{table:enso_scores} compares the performance of our Earthformer with baselines on the ICAR-ENSO dataset. 
We report the mean correlation skill $C\mbox{-}\mathtt{Nino3.4}\mbox{-}\mathtt{M}=\frac{1}{K}\sum_k \Cninok{}$ and the weighted mean correlation skill $C\mbox{-}\mathtt{Nino3.4}\mbox{-}\mathtt{WM}=\frac{1}{K}\sum_k a_k\cdot \Cninok{}$ over $K=12$ forecasting steps\footnote{$a_k = b_k\cdot\ln k$, where $b_k = 1.5\text{,\ for}\ k\leq4$; $b_k = 2\text{,\ for}\ 4<k\leq11$; $b_k = 3\text{,\ for}\ \ k > 11.$}, as well as the MSE between the spatiotemporal SST anomalies prediction and the corresponding ground-truth. 
We can find that Earthformer consistently outperforms the baselines in all concerned evaluation metrics and that using global vectors further improves the performance.

\begin{table}[!tb]
\vskip -0.7cm
\caption{Performance comparison on ICAR-ENSO. $C\mbox{-}\mathtt{Nino3.4}\mbox{-}\mathtt{M}$ and $C\mbox{-}\mathtt{Nino3.4}\mbox{-}\mathtt{WM}$ are the mean and the weighted mean of the correlation skill $\Cnino{}$ over $K=12$ forecasting steps. $C\mbox{-}\mathtt{Nino3.4}\mbox{-}\mathtt{WM}$ assigns more weights to longer-term prediction scores. MSE is calculated between the spatiotemporal SST anomalies prediction and the corresponding ground-truth.}
\label{table:enso_scores}
	\begin{center}
	\resizebox{0.7\textwidth}{!}{
	\begin{tabular}{l|c|c|ccc}
	\toprule[1.5pt]
	\multirow{2}{*}{Model}              & \multirow{2}{*}{\#Param. (M)}& \multirow{2}{*}{GFLOPS}  & \multicolumn{3}{|c}{Metrics}\\
 						                &           &		    & $C\mbox{-}\mathtt{Nino3.4}\mbox{-}\mathtt{M}\uparrow$& $C\mbox{-}\mathtt{Nino3.4}\mbox{-}\mathtt{WM}\uparrow$ & MSE ($10^{-4}$) $\downarrow$\\
	\midrule\midrule
	Persistence			                & -         & -		    & 0.3221				& 0.447				& 4.581				\\
	UNet~\cite{veillette2020sevir}      & 12.1		&\ \ 0.4    & 0.6926				& 2.102				& 2.868				\\
	ConvLSTM~\cite{shi2015convolutional}& 14.0      & 11.1      & 0.6955				& 2.107				& 2.657				\\
	PredRNN~\cite{wang2022predrnn}      & 23.8      & 85.8	    & 0.6492				& 1.910				& 3.044				\\
	PhyDNet~\cite{guen2020disentangling}&\ \ 3.1    &\ \ 5.7    & 0.6646				& 1.965				& 2.708				\\
	E3D-LSTM~\cite{wang2018eidetic}     & 12.9      & 99.8	    & 0.7040				& 2.125				& 3.095				\\
	Rainformer~\cite{bai2022rainformer} & 19.2      &\ \ 1.3    & 0.7106				& 2.153				& 3.043				\\
	\midrule
	Earthformer w/o global              &\ \ 6.6    & 23.6	    & \underline{0.7239}	& \underline{2.214}	& \underline{2.550}	\\
	Earthformer			                &\ \ 7.6    & 23.9	    & \textbf{0.7329}		& \textbf{2.259}	& \textbf{2.546}	\\
	\bottomrule[1.5pt]
	\end{tabular}
	}  
	\end{center}
	\vspace{-1.5em}
\end{table}

\section{Conclusions and Broader Impact}
\label{sec:conclusion}
In this paper, we propose Earthformer, a space-time Transformer for Earth system forecasting. Earthformer is based on a generic and efficient building block called \emph{Cuboid Attention}. It achieves SOTA on MovingMNIST, our newly proposed \nbody{}, SEVIR, and ICAR-ENSO.

Our work has certain limitations.
The first one is that Earthformer is a deterministic model that does not model uncertainty. This may result in predicting the average of all plausible futures, causing the model to generate blurry predictions of low perceptual quality and be lack of valuable small-scale details. In fact, the community lacks appropriate metrics that measure the uncertainty component in Earth system forecasting models. Extending Earthformer to a probabilistic forecasting model can be an exciting future direction.
We include more detailed discussions and preliminary experiments about handling uncertainty in Appendix~\ref{sec:nar-vs-ar}.
The second one is that the model is purely data-driven and does not take advantage of the physical knowledge of the Earth system. Recent studies on adding physical constraints~\cite{krishnapriyan2021characterizing,negiar2022learning} and ensembling the predictions from a data-driven model and a physics-based model~\cite{ravuri2021skilful} imply that it is an active and promising research direction to pursue. We plan to study how to incorporate physical knowledge into Earthformer in the future.



\begin{ack}
This work has been made possible by a Research Impact Fund project (R6003-21) and an Innovation and Technology Fund project (ITS/004/21FP) funded by the Hong Kong Government.
\end{ack}

\bibliographystyle{plain}
\bibliography{earthformer.bib}

\begin{thebibliography}{10}

\bibitem{andersson2021seasonal}
Tom~R Andersson, J~Scott Hosking, Mar{\'\i}a P{\'e}rez-Ortiz, Brooks Paige,
  Andrew Elliott, Chris Russell, Stephen Law, Daniel~C Jones, Jeremy Wilkinson,
  Tony Phillips, et~al.
\newblock Seasonal {Arctic} sea ice forecasting with probabilistic deep
  learning.
\newblock {\em Nature communications}, 12(1):1--12, 2021.

\bibitem{arnab2021vivit}
Anurag Arnab, Mostafa Dehghani, Georg Heigold, Chen Sun, Mario Lu{\v{c}}i{\'c},
  and Cordelia Schmid.
\newblock {ViViT}: A video vision transformer.
\newblock In {\em International Conference on Computer Vision (ICCV)}, 2021.

\bibitem{bai2022rainformer}
Cong Bai, Feng Sun, Jinglin Zhang, Yi~Song, and Shengyong Chen.
\newblock Rainformer: Features extraction balanced network for radar-based
  precipitation nowcasting.
\newblock {\em IEEE Geoscience and Remote Sensing Letters}, 19:1--5, 2022.

\bibitem{bertasius2021space}
Gedas Bertasius, Heng Wang, and Lorenzo Torresani.
\newblock Is space-time attention all you need for video understanding.
\newblock {\em arXiv preprint arXiv:2102.05095}, 2(3):4, 2021.

\bibitem{chen2021glit}
Boyu Chen, Peixia Li, Chuming Li, Baopu Li, Lei Bai, Chen Lin, Ming Sun, Junjie
  Yan, and Wanli Ouyang.
\newblock {GLiT}: Neural architecture search for global and local image
  transformer.
\newblock In {\em Proceedings of the IEEE/CVF International Conference on
  Computer Vision}, pages 12--21, 2021.

\bibitem{de2020rainbench}
Christian~Schroeder de~Witt, Catherine Tong, Valentina Zantedeschi, Daniele
  De~Martini, Freddie Kalaitzis, Matthew Chantry, Duncan Watson-Parris, and
  Piotr Bilinski.
\newblock {RainBench}: towards global precipitation forecasting from satellite
  imagery.
\newblock In {\em AAAI}, 2021.

\bibitem{devlin2018bert}
Jacob Devlin, Ming-Wei Chang, Kenton Lee, and Kristina Toutanova.
\newblock {BERT}: Pre-training of deep bidirectional transformers for language
  understanding.
\newblock {\em arXiv preprint arXiv:1810.04805}, 2018.

\bibitem{dosovitskiy2020image}
Alexey Dosovitskiy, Lucas Beyer, Alexander Kolesnikov, Dirk Weissenborn,
  Xiaohua Zhai, Thomas Unterthiner, Mostafa Dehghani, Matthias Minderer, Georg
  Heigold, Sylvain Gelly, et~al.
\newblock An image is worth 16x16 words: Transformers for image recognition at
  scale.
\newblock {\em arXiv preprint arXiv:2010.11929}, 2020.

\bibitem{espeholt2021skillful}
Lasse Espeholt, Shreya Agrawal, Casper S{\o}nderby, Manoj Kumar, Jonathan Heek,
  Carla Bromberg, Cenk Gazen, Jason Hickey, Aaron Bell, and Nal Kalchbrenner.
\newblock Skillful twelve hour precipitation forecasts using large context
  neural networks.
\newblock {\em arXiv preprint arXiv:2111.07470}, 2021.

\bibitem{eyring2016overview}
Veronika Eyring, Sandrine Bony, Gerald~A Meehl, Catherine~A Senior, Bjorn
  Stevens, Ronald~J Stouffer, and Karl~E Taylor.
\newblock Overview of the coupled model intercomparison project phase 6
  ({CMIP6}) experimental design and organization.
\newblock {\em Geoscientific Model Development}, 9(5):1937--1958, 2016.

\bibitem{fan2021multiscale}
Haoqi Fan, Bo~Xiong, Karttikeya Mangalam, Yanghao Li, Zhicheng Yan, Jitendra
  Malik, and Christoph Feichtenhofer.
\newblock Multiscale vision transformers.
\newblock In {\em ICCV}, 2021.

\bibitem{goodman2019goes}
Steven~J Goodman, Timothy~J Schmit, Jaime Daniels, and Robert~J Redmon.
\newblock {\em The {GOES-R} series: a new generation of geostationary
  environmental satellites}.
\newblock Elsevier, 2019.

\bibitem{guen2020disentangling}
Vincent~Le Guen and Nicolas Thome.
\newblock Disentangling physical dynamics from unknown factors for unsupervised
  video prediction.
\newblock In {\em Proceedings of the IEEE/CVF Conference on Computer Vision and
  Pattern Recognition}, pages 11474--11484, 2020.

\bibitem{guibas2021adaptive}
John Guibas, Morteza Mardani, Zongyi Li, Andrew Tao, Anima Anandkumar, and
  Bryan Catanzaro.
\newblock Adaptive fourier neural operators: Efficient token mixers for
  transformers.
\newblock {\em arXiv preprint arXiv:2111.13587}, 2021.

\bibitem{ham2019deep}
Yoo-Geun Ham, Jeong-Hwan Kim, and Jing-Jia Luo.
\newblock Deep learning for multi-year {ENSO} forecasts.
\newblock {\em Nature}, 573(7775):568--572, 2019.

\bibitem{TNT}
Kai Han, An~Xiao, Enhua Wu, Jianyuan Guo, Chunjing Xu, and Yunhe Wang.
\newblock Transformer in transformer.
\newblock {\em NeurIPS}, 2021.

\bibitem{heiss1990nexrad}
William~H Heiss, David~L McGrew, and Dale Sirmans.
\newblock {NEXRAD}: next generation weather radar (wsr-88d).
\newblock {\em Microwave Journal}, 33(1):79--89, 1990.

\bibitem{hendrycks2016gaussian}
Dan Hendrycks and Kevin Gimpel.
\newblock Gaussian error linear units (gelus).
\newblock {\em arXiv preprint arXiv:1606.08415}, 2016.

\bibitem{ho2019axial}
Jonathan Ho, Nal Kalchbrenner, Dirk Weissenborn, and Tim Salimans.
\newblock Axial attention in multidimensional transformers.
\newblock {\em arXiv preprint arXiv:1912.12180}, 2019.

\bibitem{krishnapriyan2021characterizing}
Aditi Krishnapriyan, Amir Gholami, Shandian Zhe, Robert Kirby, and Michael~W
  Mahoney.
\newblock Characterizing possible failure modes in physics-informed neural
  networks.
\newblock {\em Advances in Neural Information Processing Systems},
  34:26548--26560, 2021.

\bibitem{letellier2019chaos}
Christophe Letellier.
\newblock {\em Chaos in nature}, volume~94.
\newblock World Scientific, 2019.

\bibitem{liu2021swin}
Ze~Liu, Yutong Lin, Yue Cao, Han Hu, Yixuan Wei, Zheng Zhang, Stephen Lin, and
  Baining Guo.
\newblock Swin transformer: Hierarchical vision transformer using shifted
  windows.
\newblock In {\em Proceedings of the IEEE/CVF International Conference on
  Computer Vision}, pages 10012--10022, 2021.

\bibitem{liu2021video}
Ze~Liu, Jia Ning, Yue Cao, Yixuan Wei, Zheng Zhang, Stephen Lin, and Han Hu.
\newblock Video swin transformer.
\newblock {\em arXiv preprint arXiv:2106.13230}, 2021.

\bibitem{lucic2018gans}
Mario Lucic, Karol Kurach, Marcin Michalski, Sylvain Gelly, and Olivier
  Bousquet.
\newblock Are {GAN}s created equal? a large-scale study.
\newblock In {\em NeurIPS}, 2018.

\bibitem{mj2006three}
Valtonen MJ, Mauri Valtonen, and Hannu Karttunen.
\newblock {\em The three-body problem}.
\newblock Cambridge University Press, 2006.

\bibitem{negiar2022learning}
Geoffrey N{\'e}giar, Michael~W Mahoney, and Aditi~S Krishnapriyan.
\newblock Learning differentiable solvers for systems with hard constraints.
\newblock {\em arXiv preprint arXiv:2207.08675}, 2022.

\bibitem{neimark2021video}
Daniel Neimark, Omri Bar, Maya Zohar, and Dotan Asselmann.
\newblock Video transformer network.
\newblock {\em arXiv preprint arXiv:2102.00719}, 2021.

\bibitem{pathak2022fourcastnet}
Jaideep Pathak, Shashank Subramanian, Peter Harrington, Sanjeev Raja, Ashesh
  Chattopadhyay, Morteza Mardani, Thorsten Kurth, David Hall, Zongyi Li, Kamyar
  Azizzadenesheli, et~al.
\newblock {FourCastNet}: A global data-driven high-resolution weather model
  using adaptive fourier neural operators.
\newblock {\em arXiv preprint arXiv:2202.11214}, 2022.

\bibitem{patrick2021keeping}
Mandela Patrick, Dylan Campbell, Yuki~M. Asano, Ishan Misra~Florian Metze,
  Christoph Feichtenhofer, Andrea Vedaldi, and Jo{\~a}o~F. Henriques.
\newblock Keeping your eye on the ball: Trajectory attention in video
  transformers.
\newblock In {\em Advances in Neural Information Processing Systems (NeurIPS)},
  2021.

\bibitem{rakhimov2020latent}
Ruslan Rakhimov, Denis Volkhonskiy, Alexey Artemov, Denis Zorin, and Evgeny
  Burnaev.
\newblock Latent video transformer.
\newblock {\em arXiv preprint arXiv:2006.10704}, 2020.

\bibitem{ramesh2021zero}
Aditya Ramesh, Mikhail Pavlov, Gabriel Goh, Scott Gray, Chelsea Voss, Alec
  Radford, Mark Chen, and Ilya Sutskever.
\newblock Zero-shot text-to-image generation.
\newblock In {\em International Conference on Machine Learning}, pages
  8821--8831. PMLR, 2021.

\bibitem{ravuri2021skilful}
Suman Ravuri, Karel Lenc, Matthew Willson, Dmitry Kangin, Remi Lam, Piotr
  Mirowski, Megan Fitzsimons, Maria Athanassiadou, Sheleem Kashem, Sam Madge,
  et~al.
\newblock Skilful precipitation nowcasting using deep generative models of
  radar.
\newblock {\em Nature}, 597(7878):672--677, 2021.

\bibitem{razavi2019generating}
Ali Razavi, Aaron Van~den Oord, and Oriol Vinyals.
\newblock Generating diverse high-fidelity images with {VQ-VAE-2}.
\newblock {\em Advances in neural information processing systems}, 32, 2019.

\bibitem{reichstein2019deep}
Markus Reichstein, Gustau Camps-Valls, Bjorn Stevens, Martin Jung, Joachim
  Denzler, Nuno Carvalhais, et~al.
\newblock Deep learning and process understanding for data-driven earth system
  science.
\newblock {\em Nature}, 566(7743):195--204, 2019.

\bibitem{Salimans2017PixeCNN}
Tim Salimans, Andrej Karpathy, Xi~Chen, and Diederik~P. Kingma.
\newblock {PixelCNN++}: A {PixelCNN} implementation with discretized logistic
  mixture likelihood and other modifications.
\newblock In {\em ICLR}, 2017.

\bibitem{shi2015convolutional}
Xingjian Shi, Zhourong Chen, Hao Wang, Dit-Yan Yeung, Wai-Kin Wong, and
  Wang-chun Woo.
\newblock Convolutional {LSTM} network: A machine learning approach for
  precipitation nowcasting.
\newblock In {\em NeurIPS}, volume~28, 2015.

\bibitem{shi2017deep}
Xingjian Shi, Zhihan Gao, Leonard Lausen, Hao Wang, Dit-Yan Yeung, Wai-kin
  Wong, and Wang-chun Woo.
\newblock Deep learning for precipitation nowcasting: A benchmark and a new
  model.
\newblock In {\em NeurIPS}, volume~30, 2017.

\bibitem{srivastava2015unsupervised}
Nitish Srivastava, Elman Mansimov, and Ruslan Salakhudinov.
\newblock Unsupervised learning of video representations using {LSTM}s.
\newblock In {\em ICML}, pages 843--852. PMLR, 2015.

\bibitem{tapley2004grace}
Byron~D Tapley, Srinivas Bettadpur, John~C Ries, Paul~F Thompson, and Michael~M
  Watkins.
\newblock {GRACE} measurements of mass variability in the earth system.
\newblock {\em science}, 305(5683):503--505, 2004.

\bibitem{van2017neural}
Aaron Van Den~Oord, Oriol Vinyals, et~al.
\newblock Neural discrete representation learning.
\newblock {\em Advances in neural information processing systems}, 30, 2017.

\bibitem{vaswani2021scaling}
Ashish Vaswani, Prajit Ramachandran, Aravind Srinivas, Niki Parmar, Blake
  Hechtman, and Jonathon Shlens.
\newblock Scaling local self-attention for parameter efficient visual
  backbones.
\newblock In {\em CVPR}, 2021.

\bibitem{vaswani2017attention}
Ashish Vaswani, Noam Shazeer, Niki Parmar, Jakob Uszkoreit, Llion Jones,
  Aidan~N Gomez, {\L}ukasz Kaiser, and Illia Polosukhin.
\newblock Attention is all you need.
\newblock In {\em NeurIPS}, volume~30, 2017.

\bibitem{veillette2020sevir}
Mark Veillette, Siddharth Samsi, and Chris Mattioli.
\newblock {SEVIR}: A storm event imagery dataset for deep learning applications
  in radar and satellite meteorology.
\newblock {\em Advances in Neural Information Processing Systems},
  33:22009--22019, 2020.

\bibitem{wang2018eidetic}
Yunbo Wang, Lu~Jiang, Ming-Hsuan Yang, Li-Jia Li, Mingsheng Long, and
  Li~Fei-Fei.
\newblock Eidetic {3D} {LSTM}: A model for video prediction and beyond.
\newblock In {\em International conference on learning representations}, 2018.

\bibitem{wang2022predrnn}
Yunbo Wang, Haixu Wu, Jianjin Zhang, Zhifeng Gao, Jianmin Wang, Philip Yu, and
  Mingsheng Long.
\newblock {PredRNN}: A recurrent neural network for spatiotemporal predictive
  learning.
\newblock {\em IEEE Transactions on Pattern Analysis and Machine Intelligence},
  2022.

\bibitem{weissenborn2019scaling}
Dirk Weissenborn, Oscar T{\"a}ckstr{\"o}m, and Jakob Uszkoreit.
\newblock Scaling autoregressive video models.
\newblock In {\em International Conference on Learning Representations}, 2019.

\bibitem{wu2018group}
Yuxin Wu and Kaiming He.
\newblock Group normalization.
\newblock In {\em Proceedings of the European conference on computer vision
  (ECCV)}, pages 3--19, 2018.

\bibitem{xiong2020layer}
Ruibin Xiong, Yunchang Yang, Di~He, Kai Zheng, Shuxin Zheng, Chen Xing,
  Huishuai Zhang, Yanyan Lan, Liwei Wang, and Tieyan Liu.
\newblock On layer normalization in the transformer architecture.
\newblock In {\em International Conference on Machine Learning}, pages
  10524--10533. PMLR, 2020.

\bibitem{yan2022multiview}
Shen Yan, Xuehan Xiong, Anurag Arnab, Zhichao Lu, Mi~Zhang, Chen Sun, and
  Cordelia Schmid.
\newblock Multiview transformers for video recognition.
\newblock In {\em CVPR}, 2022.

\bibitem{yan2021videogpt}
Wilson Yan, Yunzhi Zhang, Pieter Abbeel, and Aravind Srinivas.
\newblock {VideoGPT}: Video generation using vq-vae and transformers.
\newblock {\em arXiv preprint arXiv:2104.10157}, 2021.

\bibitem{yang2021focal}
Jianwei Yang, Chunyuan Li, Pengchuan Zhang, Xiyang Dai, Bin Xiao, Lu~Yuan, and
  Jianfeng Gao.
\newblock Focal self-attention for local-global interactions in vision
  transformers.
\newblock In {\em NeurIPS}, 2021.

\bibitem{ying2021transformers}
Chengxuan Ying, Tianle Cai, Shengjie Luo, Shuxin Zheng, Guolin Ke, Di~He,
  Yanming Shen, and Tie-Yan Liu.
\newblock Do transformers really perform badly for graph representation?
\newblock {\em Advances in Neural Information Processing Systems}, 34, 2021.

\bibitem{zaheer2020big}
Manzil Zaheer, Guru Guruganesh, Kumar~Avinava Dubey, Joshua Ainslie, Chris
  Alberti, Santiago Ontanon, Philip Pham, Anirudh Ravula, Qifan Wang, Li~Yang,
  et~al.
\newblock Big bird: Transformers for longer sequences.
\newblock {\em Advances in Neural Information Processing Systems},
  33:17283--17297, 2020.

\bibitem{zhang2021vidtr}
Yanyi Zhang, Xinyu Li, Chunhui Liu, Bing Shuai, Yi~Zhu, Biagio Brattoli, Hao
  Chen, Ivan Marsic, and Joseph Tighe.
\newblock Vidtr: Video transformer without convolutions.
\newblock In {\em Proceedings of the IEEE/CVF International Conference on
  Computer Vision}, pages 13577--13587, 2021.

\end{thebibliography}

\newpage
\section*{Checklist}


\begin{enumerate}
\item For all authors...
\begin{enumerate}
  \item Do the main claims made in the abstract and introduction accurately reflect the paper's contributions and scope?
    \answerYes{}
  \item Did you describe the limitations of your work?
    \answerYes{See Section~\ref{sec:conclusion}.}
  \item Did you discuss any potential negative societal impacts of your work?
    \answerNo{Exploring Earth system forecasting has no negative societal impacts.}
  \item Have you read the ethics review guidelines and ensured that your paper conforms to them?
    \answerYes{}
\end{enumerate}

\item If you are including theoretical results...
\begin{enumerate}
  \item Did you state the full set of assumptions of all theoretical results?
    \answerNA{}
        \item Did you include complete proofs of all theoretical results?
    \answerNA{}
\end{enumerate}

\item If you ran experiments...
\begin{enumerate}
  \item Did you include the code, data, and instructions needed to reproduce the main experimental results (either in the supplemental material or as a URL)?
    \answerYes{}
  \item Did you specify all the training details (e.g., data splits, hyperparameters, how they were chosen)?
    \answerYes{}
        \item Did you report error bars (e.g., with respect to the random seed after running experiments multiple times)?
    \answerNo{}
        \item Did you include the total amount of compute and the type of resources used (e.g., type of GPUs, internal cluster, or cloud provider)?
    \answerYes{}
\end{enumerate}

\item If you are using existing assets (e.g., code, data, models) or curating/releasing new assets...
\begin{enumerate}
  \item If your work uses existing assets, did you cite the creators?
    \answerYes{}
  \item Did you mention the license of the assets?
    \answerYes{}
  \item Did you include any new assets either in the supplemental material or as a URL?
    \answerNo{}
  \item Did you discuss whether and how consent was obtained from people whose data you're using/curating?
    \answerNA{}
  \item Did you discuss whether the data you are using/curating contains personally identifiable information or offensive content?
    \answerNA{}
\end{enumerate}

\item If you used crowdsourcing or conducted research with human subjects...
\begin{enumerate}
  \item Did you include the full text of instructions given to participants and screenshots, if applicable?
    \answerNA{}
  \item Did you describe any potential participant risks, with links to Institutional Review Board (IRB) approvals, if applicable?
    \answerNA{}
  \item Did you include the estimated hourly wage paid to participants and the total amount spent on participant compensation?
    \answerNA{}
\end{enumerate}

\end{enumerate}

\clearpage
\appendix

\section{Implementation Details}
\label{sec:implementatoin-details}
All experiments are conducted on machines with NVIDIA V100 GPUs. All models including Earthformer and the baselines can fit in a single GPU (with gradient checkpointing) and get trained without model parallelization.

\subsection{Earthformer}
 \paragraph{Cuboid attention block}
 A cuboid attention block consists of $L$ cuboid attention layers with different hyperparameters, as shown in~\ref{table:cuboid_attention_patterns}. For example, as illustrated in the ``Configuration Values'' of~\tabref{table:cuboid_attention_patterns}, ``Axial'' has $L=3$ since it has $3$ consecutive patterns $\mathtt{(T, 1, 1)\rightarrow (1, H, 1)\rightarrow (1, 1, W)}$, ``Divided Space-Time'' and ``Video Swin $P\times M$'' have $L=2$. We also add Layer Normalization (LN) and Feed-Forward Networks (FFNs) in the same way as Pre-LN Transformer~\cite{xiong2020layer}. Given the input tensor $\mathcal{X}^{d-1}$ and global vectors $\mathcal{G}^{d-1}$, the output $(\mathcal{X}^{d}, \mathcal{G}^{d})$ of the $d$-th cuboid attention block (there are $D$ cuboid attention blocks in total for each hierarchy as illustrated in~\figref{fig:enc_dec}), is computed as~\eqnref{eq:cuboid_attention_block_compute}:

 \begin{align}
    \centering
    \begin{aligned}
        \mathcal{X}^{d,1} &= \mathtt{FFN}_{d,1}^{\mathtt{local}}\left(\mathcal{X}^{d-1}+\mathtt{CubAttn}_{d,1}^{\mathtt{local}}\left(\mathtt{LN}_{d,1}^{\mathtt{local}}(\mathcal{X}^{d-1}),\mathtt{LN}_{d,1}^{\mathtt{global}}(\mathcal{G}^{d-1})\right)\right), \\
        \mathcal{G}^{d,1} &= \mathtt{FFN}_{d,1}^{\mathtt{global}}\left(\mathcal{G}^{d-1}+\mathtt{Attn}_{d,1}^{\mathtt{global}}\left(\mathtt{LN}_{d,1}^{\mathtt{global}}(\mathcal{G}^{d-1}), \mathtt{LN}_{d,1}^{\mathtt{local}}(\mathcal{X}^{d-1})\right)\right), \\
        &\;\;\vdots \\
        \mathcal{X}^{d,l} &= \mathtt{FFN}_{d,l}^{\mathtt{local}}\left(\mathcal{X}^{d,l-1}+\mathtt{CubAttn}_{d,l}^{\mathtt{local}}\left(\mathtt{LN}_{d,l}^{\mathtt{local}}(\mathcal{X}^{d,l-1}), \mathtt{LN}_{d,l}^{\mathtt{global}}(\mathcal{G}^{d,l-1})\right)\right), \\
        \mathcal{G}^{d,l} &= \mathtt{FFN}_{d,l}^{\mathtt{global}}\left(\mathcal{G}^{d,l-1}+\mathtt{Attn}_{d,l}^{\mathtt{global}}\left(\mathtt{LN}_{d,l}^{\mathtt{global}}(\mathcal{G}^{d,l-1}), \mathtt{LN}_{d,l}^{\mathtt{local}}(\mathcal{X}^{d,l-1})\right)\right), \\
        &\;\;\vdots \\
        \mathcal{X}^{d}=\mathcal{X}^{d,L} &= \mathtt{FFN}_{d,L}^{\mathtt{local}}\left(\mathcal{X}^{d,L-1}+\mathtt{CubAttn}_{d,L}^{\mathtt{local}}\left(\mathtt{LN}_{d,L}^{\mathtt{local}}(\mathcal{X}^{d,L-1}),\mathtt{LN}_{d,L}^{\mathtt{global}}(\mathcal{G}^{d,L-1})\right)\right), \\
        \mathcal{G}^{d}=\mathcal{G}^{d,L} &= \mathtt{FFN}_{d,L}^{\mathtt{global}}\left(\mathcal{G}^{d,L-1}+\mathtt{Attn}_{d,L}^{\mathtt{global}}\left(\mathtt{LN}_{d,L}^{\mathtt{global}}(\mathcal{G}^{d,L-1}), \mathtt{LN}_{d,L}^{\mathtt{local}}(\mathcal{X}^{d,L-1})\right)\right). \\
    \end{aligned}
    \label{eq:cuboid_attention_block_compute}
 \end{align}
 
 Here, $\mathtt{FFN}_{d,l}^{\mathtt{local}}, \mathtt{LN}_{d,l}^{\mathtt{local}}$ means the FFN an LN layers applied on the tensor $\mathcal{X}^{d,l-1}$, and $\mathtt{FFN}_{d,l}^{\mathtt{global}}, \mathtt{LN}_{d,l}^{\mathtt{global}}$ means those applied on the global vectors $\mathcal{G}^{d,l-1}$. 
 $\mathtt{CubAttn}_{d,l}^{\mathtt{local}}$ is the combination of ``decompose'', ``attend'' and ``merge'' pipeline described in~\eqnref{eq:cuboid_decompose},\ref{eq:cuboid_l2g_attention},\ref{eq:cuboid_merge}.
 We omit cuboid\_size, strategy and shift for brevity.
 $\mathtt{Attn}_{d,l}^{\mathtt{global}}$ is equivalent to~\eqnref{eq:cuboid_g2l_attention}.
 For the $l$-th cuboid pattern in a cuboid attention block, the input $\mathcal{X}^{d,l-1}$ and $\mathcal{G}^{d,l-1}$ go through the LN layer, cuboid attention layer and FFN layer with residual connections sequentially, produce the output $\mathcal{X}^{d,l}$ and $\mathcal{G}^{d,l}$. 
 The output of the final cuboid pattern $\mathcal{X}^{d,L}$ and $\mathcal{G}^{d,L}$ serve as the output of the whole cuboid attention block $\mathcal{X}^{d}$ and $\mathcal{G}^{d}$.

\paragraph{Hyperparameters of Earthformer architecture}
The detailed architecture configurations of Earthformer are described in~\tabref{table:earthformer_detail_mnist} and~\tabref{table:earthformer_detail_sevir}. We adopt the same configurations for MovingMNIST, \nbody{} and ICAR-ENSO datasets, as shown in~\tabref{table:earthformer_detail_mnist}. We slightly adjust the configurations on SEVIR as shown in~\tabref{table:earthformer_detail_sevir}, due to its high resolution and large dataset size.

\begin{table}[!tb]
\caption{
The details of the Earthformer model on MovingMNIST, \nbody{} and ICAR-ENSO datasets. 
$\mathtt{Conv3\times3}$ is the 2D convolutional layer with $3\times3$ kernel. 
$\mathtt{GroupNorm16}$ is the Group Normalization (GN) layer~\cite{wu2018group} with $16$ groups. 
The negative slope in $\mathtt{LeakyReLU}$ is $0.1$. 
The $\mathtt{FFN}$ consists of two $\mathtt{Linear}$ layers separated by a $\mathtt{GeLU}$ activation layer~\cite{hendrycks2016gaussian}. 
$\mathtt{PatchMerge}$ splits a 2D input tensor with $C$ channels into $N$ non-overlapping $p\times p$ patches and merges the spatial dimensions into channels, gets $N$ $1\times1$ patches with $p^2\cdot C$ channels and concatenates them back along spatial dimensions.}
\label{table:earthformer_detail_mnist}
\begin{center}
	\resizebox{0.9\textwidth}{!}{
	\begin{tabular}{l|l|c|c}
	\toprule[1.5pt]
	Block                                               & Layer                     & Resolution                        & Channels          \\
    \midrule\midrule
    Input                                               & -                         & $64\times64$                      & $1$               \\\hline     
    \multirow{6}{*}{2D CNN+Downsampler}                 & $\mathtt{Conv3\times3}$   & $64\times64$                      & $1\rightarrow64$  \\
                                                        & $\mathtt{GroupNorm16}$    & $64\times64$                      & $64$              \\
                                                        & $\mathtt{LeakyReLU}$      & $64\times64$                      & $64$              \\
                                                        & $\mathtt{PatchMerge}$     & $64\times64\rightarrow32\times32$ & $64\rightarrow256$\\
                                                        & $\mathtt{LayerNorm}$      & $32\times32$                      & $256$             \\
                                                        & $\mathtt{Linear}$         & $32\times32$                      & $256\rightarrow64$\\\hline
    Encoder Positional Embedding                        & $\mathtt{PosEmbed}$       & $32\times32$                      & $64$              \\\hline
    \multirow{9}{*}{Cuboid Attention Block $\times 4$}  & $\mathtt{LayerNorm}$      & $32\times32$                      & $64$              \\
                                                        & $\mathtt{Cuboid(T,1,1)}$  & $32\times32$                      & $64$              \\
                                                        & $\mathtt{FFN}$            & $32\times32$                      & $64$              \\
                                                        & $\mathtt{LayerNorm}$      & $32\times32$                      & $64$              \\
                                                        & $\mathtt{Cuboid(1,H,1)}$  & $32\times32$                      & $64$              \\
                                                        & $\mathtt{FFN}$            & $32\times32$                      & $64$              \\
                                                        & $\mathtt{LayerNorm}$      & $32\times32$                      & $64$              \\
                                                        & $\mathtt{Cuboid(1,1,W)}$  & $32\times32$                      & $64$              \\
                                                        & $\mathtt{FFN}$            & $32\times32$                      & $64$              \\\hline
    \multirow{3}{*}{Downsampler}                        & $\mathtt{PatchMerge}$     & $32\times32\rightarrow16\times16$ & $64\rightarrow256$\\
                                                        & $\mathtt{LayerNorm}$      & $16\times16$                      & $256$             \\
                                                        & $\mathtt{Linear}$         & $16\times16$                      & $256\rightarrow128$\\\hline  
    \multirow{9}{*}{Cuboid Attention Block $\times 4$}  & $\mathtt{LayerNorm}$      & $16\times16$                      & $128$             \\
                                                        & $\mathtt{Cuboid(T,1,1)}$  & $16\times16$                      & $128$             \\
                                                        & $\mathtt{FFN}$            & $16\times16$                      & $128$             \\
                                                        & $\mathtt{LayerNorm}$      & $16\times16$                      & $128$             \\
                                                        & $\mathtt{Cuboid(1,H,1)}$  & $16\times16$                      & $128$             \\
                                                        & $\mathtt{FFN}$            & $16\times16$                      & $128$             \\
                                                        & $\mathtt{LayerNorm}$      & $16\times16$                      & $128$             \\
                                                        & $\mathtt{Cuboid(1,1,W)}$  & $16\times16$                      & $128$             \\
                                                        & $\mathtt{FFN}$            & $16\times16$                      & $128$             \\\hline
    \midrule
    Decoder Initial Positional Embedding                & $\mathtt{PosEmbed}$       & $16\times16$                      & $128$             \\\hline
    \multirow{12}{*}{Cuboid Cross Attention Block $\times 4$}& $\mathtt{LayerNorm}$      & $16\times16$                 & $128$             \\
                                                        & $\mathtt{Cuboid(T,1,1)}$  & $16\times16$                      & $128$             \\
                                                        & $\mathtt{FFN}$            & $16\times16$                      & $128$             \\
                                                        & $\mathtt{LayerNorm}$      & $16\times16$                      & $128$             \\
                                                        & $\mathtt{Cuboid(1,H,1)}$  & $16\times16$                      & $128$             \\
                                                        & $\mathtt{FFN}$            & $16\times16$                      & $128$             \\
                                                        & $\mathtt{LayerNorm}$      & $16\times16$                      & $128$             \\
                                                        & $\mathtt{Cuboid(1,1,W)}$  & $16\times16$                      & $128$             \\
                                                        & $\mathtt{FFN}$            & $16\times16$                      & $128$             \\
                                                        & $\mathtt{CuboidCross(T,1,1)}$  & $16\times16$                 & $128$             \\
                                                        & $\mathtt{FFN}$            & $16\times16$                      & $128$             \\
                                                        & $\mathtt{LayerNorm}$      & $16\times16$                      & $128$             \\\hline
    \multirow{2}{*}{Upsampler}                          & $\mathtt{NearestNeighborInterp}$  & $16\times16\rightarrow32\times32$ & $128$     \\
                                                        & $\mathtt{Conv3\times3}$   & $32\times32$                      & $128\rightarrow64$\\\hline 
    \multirow{12}{*}{Cuboid Cross Attention Block $\times 4$}& $\mathtt{LayerNorm}$      & $32\times32$                 & $64$              \\
                                                        & $\mathtt{Cuboid(T,1,1)}$  & $32\times32$                      & $64$              \\
                                                        & $\mathtt{FFN}$            & $32\times32$                      & $64$              \\
                                                        & $\mathtt{LayerNorm}$      & $32\times32$                      & $64$              \\
                                                        & $\mathtt{Cuboid(1,H,1)}$  & $32\times32$                      & $64$              \\
                                                        & $\mathtt{FFN}$            & $32\times32$                      & $64$              \\
                                                        & $\mathtt{LayerNorm}$      & $32\times32$                      & $64$              \\
                                                        & $\mathtt{Cuboid(1,1,W)}$  & $32\times32$                      & $64$              \\
                                                        & $\mathtt{FFN}$            & $32\times32$                      & $64$              \\
                                                        & $\mathtt{CuboidCross(T,1,1)}$  & $32\times32$                 & $64$              \\
                                                        & $\mathtt{FFN}$            & $32\times32$                      & $64$              \\
                                                        & $\mathtt{LayerNorm}$      & $32\times32$                      & $64$              \\\hline                                                
	\multirow{5}{*}{2D CNN+Upsampler}                   & $\mathtt{NearestNeighborInterp}$  & $32\times32\rightarrow64\times64$ & $64$      \\
	                                                    & $\mathtt{Conv3\times3}$   & $64\times64$                      & $64$              \\
                                                        & $\mathtt{GroupNorm16}$    & $64\times64$                      & $64$              \\
                                                        & $\mathtt{LeakyReLU}$      & $64\times64$                      & $64$              \\
                                                        & $\mathtt{Linear}$         & $64\times64$                      & $64\rightarrow1$  \\
	\bottomrule[1.5pt]
	\end{tabular}
	}  
	\end{center}
\end{table}

\begin{table}[!tb]
\caption{
The details of the Earthformer model on SEVIR dataset. $\mathtt{Conv3\times3}$ is the 2D convolutional layer with $3\times3$ kernel. 
$\mathtt{GroupNorm16}$ is the Group Normalization (GN) layer~\cite{wu2018group} with $16$ groups. 
The negative slope in $\mathtt{LeakyReLU}$ is $0.1$. 
The $\mathtt{FFN}$ consists of two $\mathtt{Linear}$ layers separated by a $\mathtt{GeLU}$ activation layer~\cite{hendrycks2016gaussian}. 
$\mathtt{PatchMerge}$ splits a 2D input tensor with $C$ channels into $N$ non-overlapping $p\times p$ patches and merges the spatial dimensions into channels, gets $N$ $1\times1$ patches with $p^2\cdot C$ channels and concatenates them back along spatial dimensions.}
\label{table:earthformer_detail_sevir}
\begin{center}
	\resizebox{0.7\textwidth}{!}{
	\begin{tabular}{l|l|c|c}
	\toprule[1.5pt]
	Block                                               & Layer                     & Resolution                        & Channels          \\
    \midrule\midrule
    Input                                               & -                         & $384\times384$                    & $1$               \\\hline     
    \multirow{7}{*}{2D CNN+Downsampler}                 & $\mathtt{Conv3\times3}$   & $384\times384$                    & $1\rightarrow16$  \\
                                                        & $\mathtt{Conv3\times3}$   & $384\times384$                    & $16$              \\
                                                        & $\mathtt{GroupNorm16}$    & $384\times384$                    & $16$              \\
                                                        & $\mathtt{LeakyReLU}$      & $384\times384$                    & $16$              \\
                                                        & $\mathtt{PatchMerge}$     & $384\times384\rightarrow128\times128$ & $16\rightarrow144$\\
                                                        & $\mathtt{LayerNorm}$      & $128\times128$                    & $144$             \\
                                                        & $\mathtt{Linear}$         & $128\times128$                    & $144\rightarrow16$\\\hline
    \multirow{7}{*}{2D CNN+Downsampler}                 & $\mathtt{Conv3\times3}$   & $128\times128$                    & $16\rightarrow64$ \\
                                                        & $\mathtt{Conv3\times3}$   & $128\times128$                    & $64$              \\
                                                        & $\mathtt{GroupNorm16}$    & $128\times128$                    & $64$              \\
                                                        & $\mathtt{LeakyReLU}$      & $128\times128$                    & $64$              \\
                                                        & $\mathtt{PatchMerge}$     & $128\times128\rightarrow64\times64$ & $64\rightarrow256$\\
                                                        & $\mathtt{LayerNorm}$      & $64\times64$                      & $256$             \\
                                                        & $\mathtt{Linear}$         & $64\times64$                      & $256\rightarrow64$\\\hline
    \multirow{7}{*}{2D CNN+Downsampler}                 & $\mathtt{Conv3\times3}$   & $64\times64$                      & $64\rightarrow128$\\
                                                        & $\mathtt{Conv3\times3}$   & $64\times64$                      & $128$             \\
                                                        & $\mathtt{GroupNorm16}$    & $64\times64$                      & $128$             \\
                                                        & $\mathtt{LeakyReLU}$      & $64\times64$                      & $128$             \\
                                                        & $\mathtt{PatchMerge}$     & $64\times64\rightarrow32\times32$ & $128\rightarrow512$\\
                                                        & $\mathtt{LayerNorm}$      & $32\times32$                      & $512$             \\
                                                        & $\mathtt{Linear}$         & $32\times32$                      & $512\rightarrow128$\\\hline
    Encoder Positional Embedding                        & $\mathtt{PosEmbed}$       & $32\times32$                      & $128$              \\\hline
    \multirow{9}{*}{Cuboid Attention Block $\times 2$}  & $\mathtt{LayerNorm}$      & $32\times32$                      & $128$              \\
                                                        & $\mathtt{Cuboid(T,1,1)}$  & $32\times32$                      & $128$              \\
                                                        & $\mathtt{FFN}$            & $32\times32$                      & $128$              \\
                                                        & $\mathtt{LayerNorm}$      & $32\times32$                      & $128$              \\
                                                        & $\mathtt{Cuboid(1,H,1)}$  & $32\times32$                      & $128$              \\
                                                        & $\mathtt{FFN}$            & $32\times32$                      & $128$              \\
                                                        & $\mathtt{LayerNorm}$      & $32\times32$                      & $128$              \\
                                                        & $\mathtt{Cuboid(1,1,W)}$  & $32\times32$                      & $128$              \\
                                                        & $\mathtt{FFN}$            & $32\times32$                      & $128$              \\\hline
    \multirow{3}{*}{Downsampler}                        & $\mathtt{PatchMerge}$     & $32\times32\rightarrow16\times16$ & $128\rightarrow512$\\
                                                        & $\mathtt{LayerNorm}$      & $16\times16$                      & $512$             \\
                                                        & $\mathtt{Linear}$         & $16\times16$                      & $512\rightarrow256$\\\hline  
    \multirow{9}{*}{Cuboid Attention Block $\times 2$}  & $\mathtt{LayerNorm}$      & $16\times16$                      & $256$             \\
                                                        & $\mathtt{Cuboid(T,1,1)}$  & $16\times16$                      & $256$             \\
                                                        & $\mathtt{FFN}$            & $16\times16$                      & $256$             \\
                                                        & $\mathtt{LayerNorm}$      & $16\times16$                      & $256$             \\
                                                        & $\mathtt{Cuboid(1,H,1)}$  & $16\times16$                      & $256$             \\
                                                        & $\mathtt{FFN}$            & $16\times16$                      & $256$             \\
                                                        & $\mathtt{LayerNorm}$      & $16\times16$                      & $256$             \\
                                                        & $\mathtt{Cuboid(1,1,W)}$  & $16\times16$                      & $256$             \\
                                                        & $\mathtt{FFN}$            & $16\times16$                      & $256$             \\\hline
    \midrule
    Decoder Initial Positional Embedding                & $\mathtt{PosEmbed}$       & $16\times16$                      & $256$             \\\hline
    \multirow{12}{*}{Cuboid Cross Attention Block $\times 2$}& $\mathtt{LayerNorm}$ & $16\times16$                      & $256$             \\
                                                        & $\mathtt{Cuboid(T,1,1)}$  & $16\times16$                      & $256$             \\
                                                        & $\mathtt{FFN}$            & $16\times16$                      & $256$             \\
                                                        & $\mathtt{LayerNorm}$      & $16\times16$                      & $256$             \\
                                                        & $\mathtt{Cuboid(1,H,1)}$  & $16\times16$                      & $256$             \\
                                                        & $\mathtt{FFN}$            & $16\times16$                      & $256$             \\
                                                        & $\mathtt{LayerNorm}$      & $16\times16$                      & $256$             \\
                                                        & $\mathtt{Cuboid(1,1,W)}$  & $16\times16$                      & $256$             \\
                                                        & $\mathtt{FFN}$            & $16\times16$                      & $256$             \\
                                                        & $\mathtt{CuboidCross(T,1,1)}$  & $16\times16$                 & $256$             \\
                                                        & $\mathtt{FFN}$            & $16\times16$                      & $256$             \\
                                                        & $\mathtt{LayerNorm}$      & $16\times16$                      & $256$             \\\hline
    \multirow{2}{*}{Upsampler}                          & $\mathtt{NearestNeighborInterp}$  & $16\times16\rightarrow32\times32$ & $256$             \\
                                                        & $\mathtt{Conv3\times3}$   & $32\times32$                      & $256\rightarrow128$\\\hline 
    \multirow{12}{*}{Cuboid Cross Attention Block $\times 2$}& $\mathtt{LayerNorm}$ & $32\times32$                      & $128$             \\
                                                        & $\mathtt{Cuboid(T,1,1)}$  & $32\times32$                      & $128$             \\
                                                        & $\mathtt{FFN}$            & $32\times32$                      & $128$             \\
                                                        & $\mathtt{LayerNorm}$      & $32\times32$                      & $128$             \\
                                                        & $\mathtt{Cuboid(1,H,1)}$  & $32\times32$                      & $128$             \\
                                                        & $\mathtt{FFN}$            & $32\times32$                      & $128$             \\
                                                        & $\mathtt{LayerNorm}$      & $32\times32$                      & $128$             \\
                                                        & $\mathtt{Cuboid(1,1,W)}$  & $32\times32$                      & $128$             \\
                                                        & $\mathtt{FFN}$            & $32\times32$                      & $128$             \\
                                                        & $\mathtt{CuboidCross(T,1,1)}$  & $32\times32$                 & $128$             \\
                                                        & $\mathtt{FFN}$            & $32\times32$                      & $128$             \\
                                                        & $\mathtt{LayerNorm}$      & $32\times32$                      & $128$             \\\hline                                                
	\multirow{4}{*}{2D CNN+Upsampler}                   & $\mathtt{NearestNeighborInterp}$  & $32\times32\rightarrow64\times64$ & $128$     \\
	                                                    & $\mathtt{Conv3\times3}$   & $64\times64$                      & $128$             \\
                                                        & $\mathtt{GroupNorm16}$    & $64\times64$                      & $128$             \\
                                                        & $\mathtt{LeakyReLU}$      & $64\times64$                      & $128$             \\\hline
    \multirow{4}{*}{2D CNN+Upsampler}                   & $\mathtt{NearestNeighborInterp}$  & $64\times64\rightarrow128\times128$ & $128$   \\
                                                        & $\mathtt{Conv3\times3}$   & $128\times128$                    & $128\rightarrow64$\\
                                                        & $\mathtt{GroupNorm16}$    & $128\times128$                    & $64$              \\
                                                        & $\mathtt{LeakyReLU}$      & $128\times128$                    & $64$              \\\hline
    \multirow{5}{*}{2D CNN+Upsampler}                   & $\mathtt{NearestNeighborInterp}$  & $128\times128\rightarrow384\times384$ & $64$  \\
                                                        & $\mathtt{Conv3\times3}$   & $384\times384$                    & $64\rightarrow16$ \\
                                                        & $\mathtt{GroupNorm16}$    & $384\times384$                    & $16$              \\
                                                        & $\mathtt{LeakyReLU}$      & $384\times384$                    & $16$              \\
                                                        & $\mathtt{Linear}$         & $384\times384$                    & $16\rightarrow1$  \\
	\bottomrule[1.5pt]
	\end{tabular}
	}  
	\end{center}
\end{table}

\paragraph{Optimization}
We train all Earthformer variants using the AdamW optimizer. Detailed configurations are shown in~\tabref{table:earthformer_optimization}. We train for $100$ epochs on all datasets and early-stop model training according to the validation score with $\mathtt{tolerance}=20$. We adopt $20\%$ linear warm-up and $\mathtt{Cosine}$ learning rate scheduler that decays the $lr$ from its maximum to zero after warm-up. We adopt data parallel and gradient accumulation to use a total batch size of $64$ while the 16GB GPU can only afford a smaller batch size like $2$ or $4$.

\begin{table}[!tb]
    \centering
    \caption{Hyperparameters of the AdamW optimizer for training Earthformer on MovingMNIST, \nbody{}, SEVIR and ICAR-ENSO datasets.}
    \begin{tabular}{l|c}
    	\toprule[1.5pt]
    	Hyper-parameter & Value \\
    	\midrule\midrule
        Learning rate               & 0.001     \\
        $\beta_1$                   & 0.9       \\
        $\beta_2$                   & 0.999     \\
        Weight decay                & 0.00001   \\
        Batch size                  & 64        \\
        Training epochs             & 100       \\
        Warm up percentage          & 20\%      \\
        Learning rate decay         & Cosine    \\
        Early stop                  & True      \\
        Early stop tolerance        & 20        \\
        \bottomrule[1.5pt]
    \end{tabular}
    \label{table:earthformer_optimization}
\end{table}

\subsection{Baselines}
We train baseline algorithms following their officially released configurations and tune the learning rate, learning rate scheduler, working resolution, etc., to optimize their performance on each dataset. We list the modifications we applied to the baselines for each dataset in~\tabref{table:baseline_impl_detail}.

\begin{table}[!tb]
\caption{Implementation details of baseline algorithms. Modifications based on the officially released implementations are listed according to different datasets. ``-'' means no modification is applied. ``reverse enc-dec'' means adopting the reversed encoder-decoder architecture proposed in~\cite{shi2017deep}. ``2D CNN downsampler/upsampler ($8\times$) wrapper'' means we wrap the model with a pair of 2D CNN downsampler and upsampler to downsample the spatial resolution $8$ times in order to reduce the GPU memory cost as well as the FLOPS. The 2D CNN downsampler and upsampler are of the same designs as those used in Earthformer. Other terms listed are the hyperparameters in their officially released implementations.}
\label{table:baseline_impl_detail}
	\begin{center}
	\resizebox{0.99\textwidth}{!}{
	\begin{tabular}{l|c|c|c|c}
	\toprule[1.5pt]
	Model                               & MovingMNIST   & \nbody{}      & SEVIR     & ICAR-ENSO \\
	\midrule\midrule
	UNet~\cite{veillette2020sevir}      & -             & -		        & -         & num\_layer=3 \\\hline
	\multirow{4}{*}{ConvLSTM~\cite{shi2015convolutional}}   & reverse enc-dec~\cite{shi2017deep} & reverse enc-dec~\cite{shi2017deep} & reverse enc-dec~\cite{shi2017deep} & reverse enc-dec~\cite{shi2017deep} \\
	& conv\_kernels = [(7,7),(5,5),(3,3)] & conv\_kernels = [(7,7),(5,5),(3,3)] & conv\_kernels = [(7,7),(5,5),(3,3)] & conv\_kernels = [(7,7),(5,5),(3,3)] \\
	& deconv\_kernels = [(6,6),(4,4),(4,4)] & deconv\_kernels = [(6,6),(4,4),(4,4)] & deconv\_kernels = [(6,6),(4,4),(4,4)] & deconv\_kernels = [(6,6),(4,4),(4,4)] \\
	& channels=[96, 128, 256] & channels=[96, 128, 256] & channels=[96, 128, 256] & channels=[96, 128, 256] \\\hline
	PredRNN~\cite{wang2022predrnn}      & -             & -		        & 2D CNN downsampler/upsampler ($8\times$) wrapper & - \\\hline
	PhyDNet~\cite{guen2020disentangling}& -             & -		        & convcell\_hidden = [256, 256, 256, 64]  & - \\\hline
	E3D-LSTM~\cite{wang2018eidetic}     & -             & -		        & 2D CNN downsampler/upsampler ($8\times$) wrapper & - \\\hline
	\multirow{5}{*}{Rainformer~\cite{bai2022rainformer}}    & downscaling\_factors=[2, 2, 2, 2] & downscaling\_factors=[2, 2, 2, 2] & downscaling\_factors=[4, 2, 2, 2] & downscaling\_factors=[1, 2, 2, 2] \\
   & hidden\_dim=32 & hidden\_dim=32    & -    & hidden\_dim=32 \\
   & heads=[4, 4, 8, 16] & heads=[4, 4, 8, 16] & - & heads=[4, 4, 8, 16] \\
   & head\_dim=8 & head\_dim=8 & - & head\_dim=8 \\
   & - & - & - & window\_size=3 \\
	\bottomrule[1.5pt]
	\end{tabular}
	}  
	\end{center}
\end{table}

\section{More Ablation Analysis on Synthetic Datasets}
As mentioned in~\secref{sec:exp_synthetic}, the design of cuboid attention greatly facilitates the search for optimal space-time attention. 
~\tabref{table:depth44_pattern_search} summarizes the cuboid attention pattern search results with the model depth equal to $4$ on both MovingMNIST and \nbody{}. 
We further investigate if it is feasible to accelerate the pattern search using shallower models. 
The results with model depth equal to $2$ are shown in~\tabref{table:depth22_pattern_search}. 
We find that the ``Axial'' pattern is still the optimal among the patterns listed in~\tabref{table:cuboid_attention_patterns}, and adding global vectors improves performance for all the patterns while having similar FLOPS. 
This observation implies that in the future work, it might be feasible to conduct NAS using smaller and shallower models with affordable cost and transfer the results to larger and deeper models.

\begin{table}[!tb]
\caption{Ablation results with depth equals to $2$. We can see that the findings still hold for shallower models.}
\label{table:depth22_pattern_search}
	\begin{center}
	\resizebox{0.99\textwidth}{!}{
	\begin{tabular}{l|ccc|ccc}
	\toprule[1.5pt]
	Model				& \multicolumn{3}{|c}{Metrics on MovingMNIST}               & \multicolumn{3}{|c}{Metrics on \nbody{}}  \\
	\midrule
 						& MSE  $\downarrow$	& MAE  $\downarrow$ & SSIM $\uparrow$   & MSE $\downarrow$	& MAE  $\downarrow$ & SSIM $\uparrow$\\
	\midrule\midrule
	Axial				& 50.41				& 106.9				& 0.8805			& 18.98				& 48.49				& 0.9391			\\
	+ global $\bigstar$	& \bgray{\textbf{49.74}}& \bgray{\underline{105.2}}& \bgray{\textbf{0.8838}}& \bgray{\textbf{18.49}}& \bgray{\underline{47.31}}& \bgray{\underline{0.9416}}	\\
	\midrule
	DST					& 56.43				& 119.5				& 0.8620			& 21.47				& 54.06				& 0.9290				\\
	+ global		    & \bgray{55.87}		& \bgray{118.2}		& \bgray{0.8635}	& \bgray{21.16}		& \bgray{53.03}		& \bgray{0.9303}	\\
	\midrule
	Video-Swin $2\times8$& \bgray{59.31}		& \bgray{124.4}		& 0.8522			& 23.04				& 56.66				& 0.9226				\\
	+ global            & 60.05				& 125.9				& \bgray{0.8522}	& \bgray{22.97}		& \bgray{55.90}		& \bgray{0.9241}				\\
	\midrule
	Video-Swin $10\times8$& 66.68				& 133.2				& \bgray{0.8387}	& 25.91				& 62.49				& 0.9100				\\
	+ global            & \bgray{65.58}		& \bgray{132.8}		& 0.8383			& \bgray{25.56}		& \bgray{61.52}		& \bgray{0.9118}				\\
	\midrule
	Spatial Local-Dilated 2& 62.85				& 131.3				& 0.8441			& 25.98				& 64.24				& 0.9075				\\
	+ global            & \bgray{60.03}		& \bgray{127.2}		& \bgray{0.8489}    & \bgray{24.90}		& \bgray{60.13}		& \bgray{0.9162}				\\
	\midrule
	Spatial Local-Dilated 4& 59.87				& 125.2				& 0.8514			& 23.68				& 57.11				& 0.9207				\\
	+ global            & \bgray{58.08}		& \bgray{120.4}		& \bgray{0.8611}	& \bgray{22.93}		& \bgray{56.22}		& \bgray{0.9236}				\\
	\midrule
	Axial Space Dilate 2& 53.12				& 110.4				& 0.8747			& 19.52				& 48.77				& 0.9378				\\
	+ global            & \bgray{51.03}		& \bgray{107.4}		& \bgray{0.8782}	& \bgray{\underline{18.53}}& \bgray{\textbf{47.00}}	& \bgray{\textbf{0.9419}}		\\
	\midrule
	Axial Space Dilate 4& 55.85				& 116.0				& 0.8645			& 20.30				& 50.51				& 0.9350			\\
	+ global            & \bgray{\underline{49.84}}& \bgray{\textbf{104.9}}	& \bgray{\underline{0.8815}}& \bgray{19.38}	& \bgray{49.32}	& \bgray{0.9380}				\\
	\bottomrule[1.5pt]
	\end{tabular}
	}  
	\end{center}
\end{table}





\clearpage
\section{Non-Auto-Regressive v.s. Auto-Regressive}
\label{sec:nar-vs-ar}
\begin{figure}[!tb]
    \centering
    \includegraphics[width=0.55\textwidth]{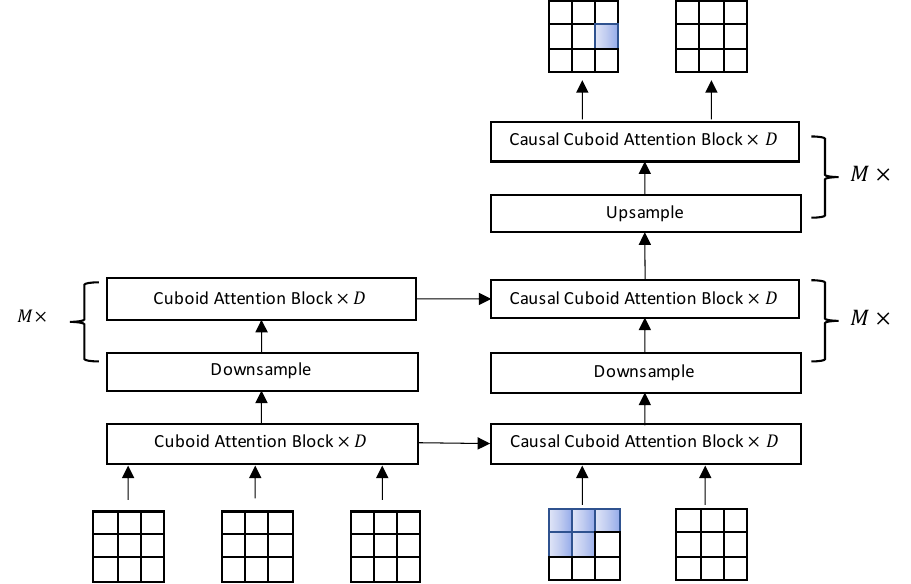}
    \caption{Illustration of the Earthformer variant in auto-regressive fashion (Earthformer AR). 
    It is a hierarchical Transformer encoder-decoder based on cuboid attention. 
    The input is a length-$3$ sequence of frames containing discrete visual codes and the target sequence has length $2$.
    ``$\times D$" means stacking $D$ (causal) cuboid attention blocks with residual connection. 
    ``$M\times$" means $M$ layers of hierarchies. 
    A causal cuboid attention block is a cuboid attention block with causal mask. 
    The ``Downsample'' and ``Upsample'' layers in the decoder follow the designs described in~\cite{Salimans2017PixeCNN} and are hence also causal.
    Causal cuboid attention blocks and causal downsampling and upsampling layers prevent each element attending to the elements with larger indices in raster-scan ordering.}
    \label{fig:enc_dec_ar}
\end{figure}

As mentioned in~\secref{sec:hierarchical_enc_dec}, besides generating the future predictions directly in a non-auto-regressive way, previous works generate the predictions patch-by-patch in raster-scan ordering~\cite{weissenborn2019scaling,yan2021videogpt,rakhimov2020latent}. 
In fact, the pros and cons of the auto-regressive and non-auto-regressive approaches are not clear under the setting of Earth system forecasting. 
Therefore, we replace the decoder of the non-auto-regressive Earthformer with the auto-regressive decoder shown in~\figref{fig:enc_dec_ar} and propose an Earthformer variant called \emph{Earthformer AR}.

The auto-regressive approach uses discrete visual tokens as the input and target. Here, we pretrain a VQ-VAE~\cite{van2017neural} with a codebook $e\in\mathbb{R}^{Q\times C^{\text{code}}}$, where $Q$ is the codebook size. The discrete latent space is hence $Q$-way categorical. 
The VQ-VAE encoder downsamples an input frame $\mathcal{X}_i\in\mathbb{R}^{H\times W\times C}$ to $\mathtt{Enc}(\mathcal{X}_i)\in\mathbb{R}^{H^\prime \times W^\prime \times C^{\text{code}}}$, then maps each element to its nearest code $e_{i}$ learned in the codebook, and produces the encoding $\mathtt{Enc}^{\text{code}}(\mathcal{X}_i)\in[Q]^{H^\prime \times W^\prime \times 1}$. 
The decoder does it reversely: it takes the latent visual codes $\mathcal{Z}^{\text{code}}_i\in[Q]^{H^\prime \times W^\prime \times 1}$, indexes the codes in the codebook to convert the discrete codes to real-valued vectors $\mathcal{Z}_i\in\mathbb{R}^{H^\prime \times W^\prime \times C^{\text{code}}}$, and then upsamples the frame back to the pixel space $\mathcal{Y}_i\in\mathbb{R}^{H\times W\times C}$.

\figref{fig:enc_dec_ar} illustrates the architecture of the Earthformer AR.
The input is a sequence of frames containing discrete visual codes $[\mathtt{Enc}^{\text{code}}(\mathcal{X}_i)]_i$, which are the output of the VQ-VAE encoder.
The target sequence is $[\mathtt{Enc}^{\text{code}}(\mathcal{Y}_i)]_i$. 
Earthformer AR generates the target code one-by-one in raster-scan ordering. 
The generation at the current step $\mathcal{Z}^{\text{code}}_{(a,b,c)}$ is conditioned on the context $\mathcal{X}$ and all the previously generated codes $[\mathcal{Z}^{\text{code}}_{<(a,b,c)}]$: 


\begin{align}
    p(\mathcal{Z}^{\text{code}}\mid\mathcal{X}) = \prod_{(a,b,c)}p\left(\mathcal{Z}^{\text{code}}_{(a,b,c)}\mid\mathcal{Z}^{\text{code}}_{<(a,b,c)}, \mathcal{X}\right).
    \label{eq:raster_scan_conditioning}
\end{align}

The VQ-VAE decoder decodes the generated visual codes $\mathcal{Z}^{\text{code}}$ to the final output in the pixel space.

\begin{figure}[!tb]
    \centering
    \includegraphics[width=0.99\textwidth]{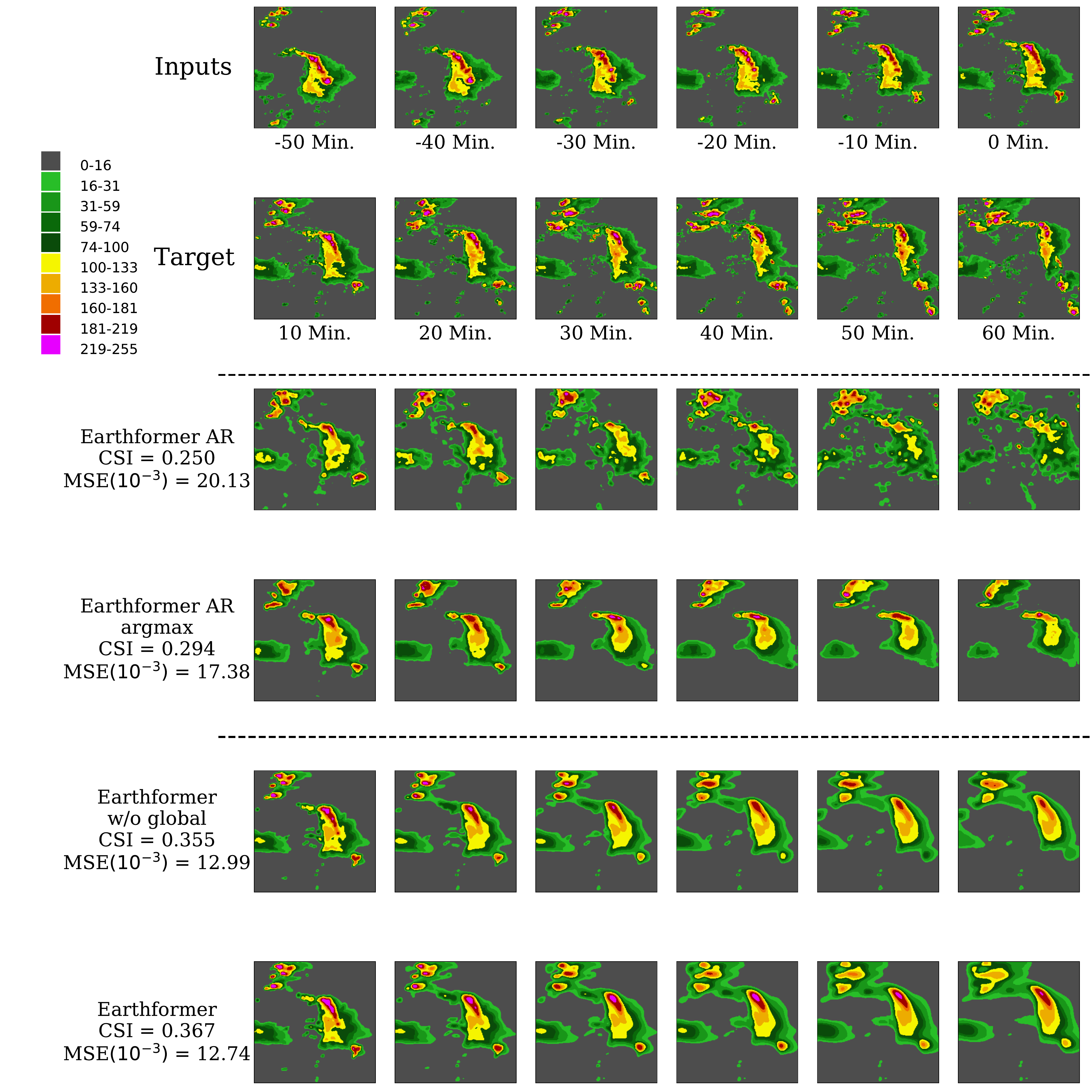}
    \caption{Non-Auto-Regressive v.s. Auto-Regressive on SEVIR: qualitative example 1.}
    \label{fig:nar_vs_ar_80}
\end{figure}

\begin{figure}[!tb]
    \centering
    \includegraphics[width=0.99\textwidth]{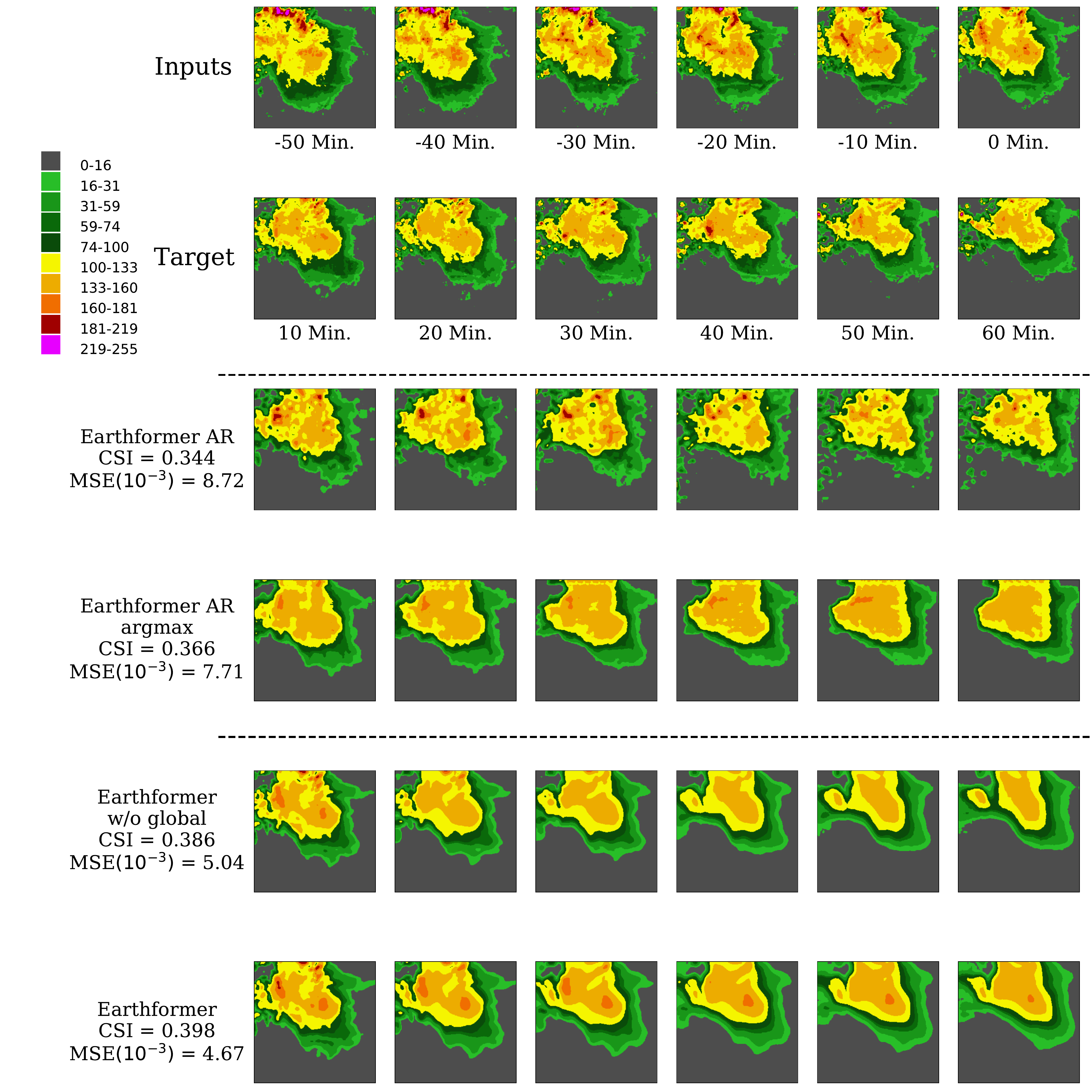}
    \caption{Non-Auto-Regressive v.s. Auto-Regressive on SEVIR: qualitative example 2.}
    \label{fig:nar_vs_ar_160}
\end{figure}

\begin{figure}[!tb]
    \centering
    \includegraphics[width=0.99\textwidth]{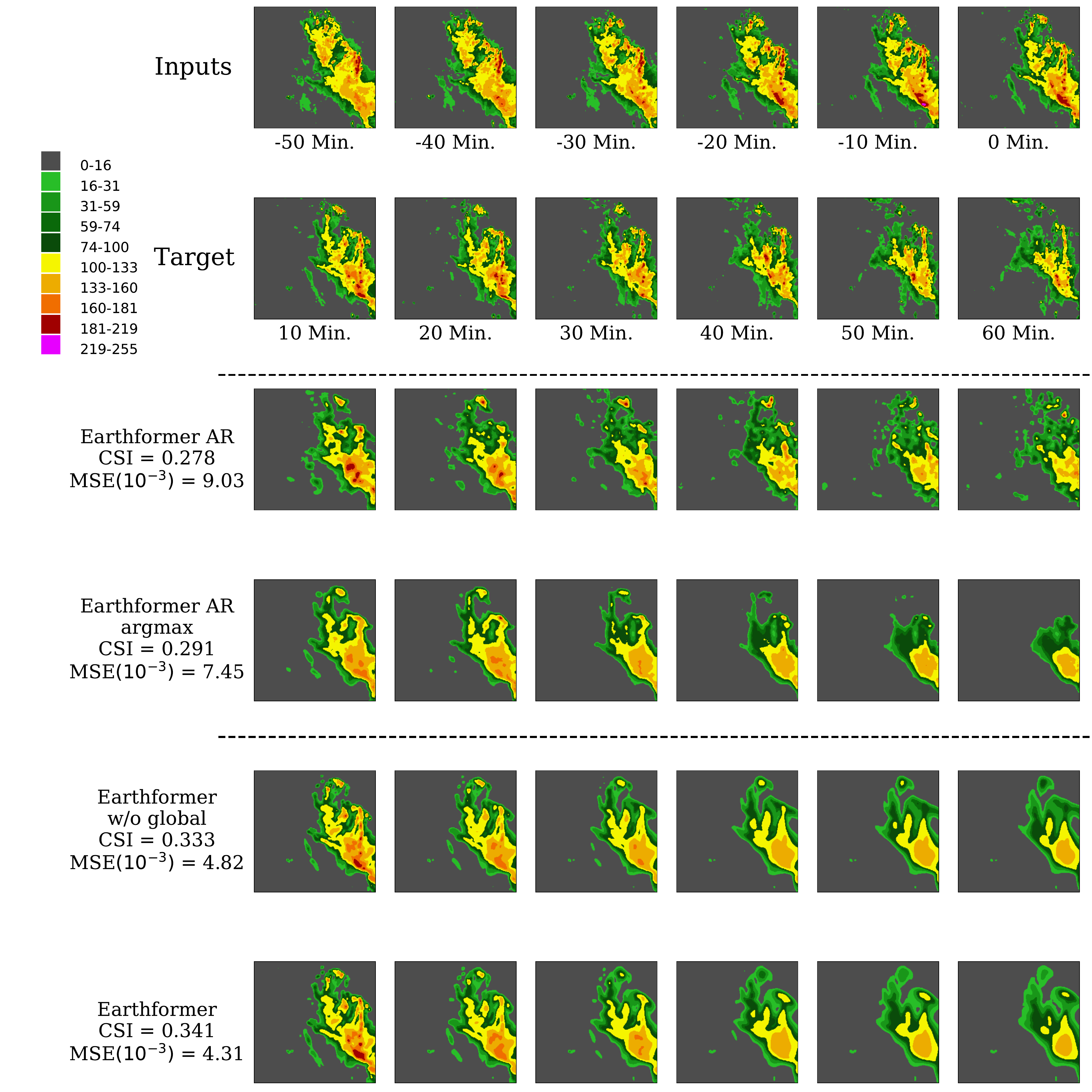}
    \caption{Non-Auto-Regressive v.s. Auto-Regressive on SEVIR: qualitative example 3.}
    \label{fig:nar_vs_ar_240}
\end{figure}

\begin{figure}[!tb]
    \centering
    \includegraphics[width=0.99\textwidth]{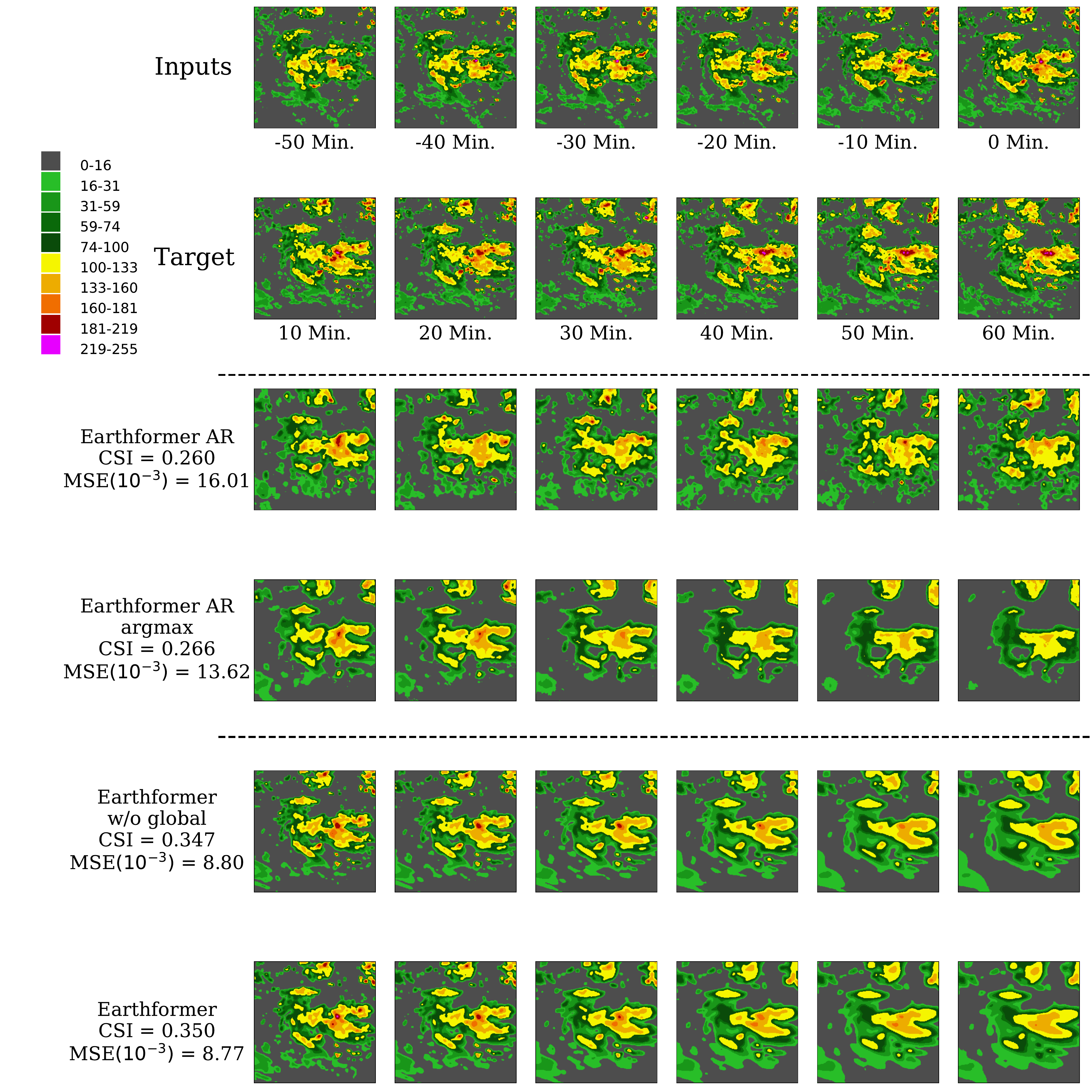}
    \caption{Non-Auto-Regressive v.s. Auto-Regressive on SEVIR: qualitative example 4.}
    \label{fig:nar_vs_ar_400}
\end{figure}

We conduct preliminary experiments of Earthformer AR on the SEVIR dataset and find that Earthformer AR is able to give more perceptually satisfying predictions than Earthformer, but is inferior in terms of skill scores. \figref{fig:nar_vs_ar_80} to~\figref{fig:nar_vs_ar_400} show the qualitative results of Earthformer and Earthformer AR on several challenging test cases. 
The outputs of Earthformer AR look more like ``real'' VIL images and do not suffer from producing blurry predictions. 
However, the performance of Earthformer AR is much worse than Earthformer (even worse than some simple baselines including UNet~\cite{veillette2020sevir}) in the concerned evaluation metrics. 
We further investigate the effect of the sampling algorithm for Earthformer AR. We compared 1) generating the codes with $\mathtt{argmax}$ step-by-step and 2) randomly drawing samples from $p\left(\mathcal{Z}^{\text{code}}_{(a,b,c)}|\mathcal{Z}^{\text{code}}_{<(a,b,c)}, \mathcal{X}_i\right)$ specified by~\eqnref{eq:raster_scan_conditioning}. We denote the $\mathtt{argmax}$-variant as \emph{Earthformer AR argmax}. We find that argmax sampling can give better skill scores but the generated results have less perceptual similarity than the ``real'' VIL images. 

Based on these results, we pick the non-auto-regressive decoder for experiments in the main paper. The fact that Earthformer AR gives more perceptually satisfying predictions than Earthformer while having worse skill scores triggers future work. So far, there is no well-established metric for evaluating the preceptual quality of the predictions generated by Earth system forecasting models. The Fréchet Inception Distance (FID) and  Inception Score (IS) adopted in evaluating image GANs~\cite{lucic2018gans} rely on a pretrained network. We are able to pretrain Earthformer on Earth observations and adopt the pretrained network for both calculating the FID and initializing the GAN discriminator.


\clearpage
\section{Chaos in N-Body MNIST Dataset}
\label{sec:chaos-nbody}
\begin{figure}[!tb]
    \centering
    \includegraphics[width=0.99\textwidth]{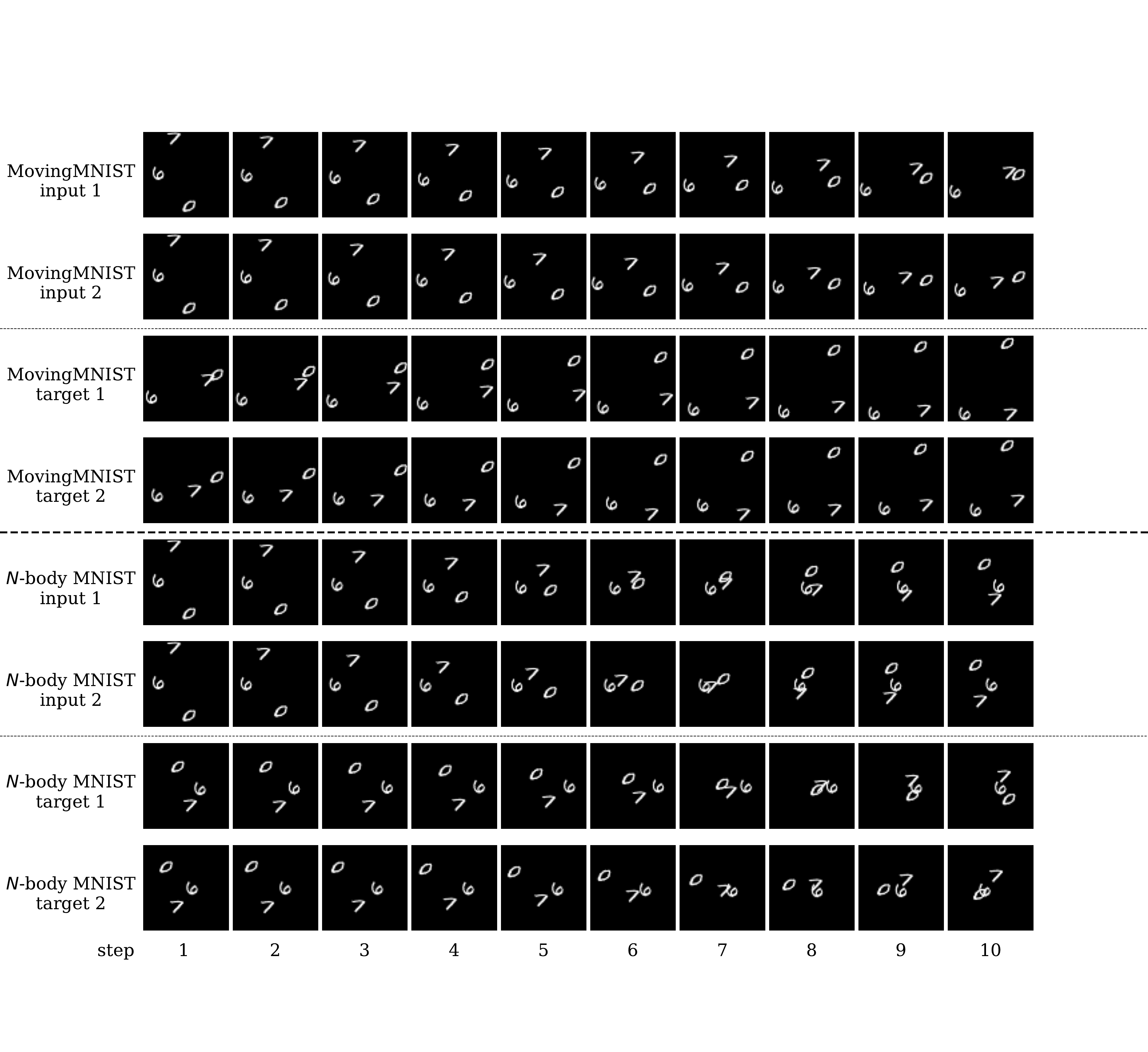}
    \caption{Chaos in \nbody{}: the effect of a slight disturbance on the initial velocities is much more significant on \nbody{} than on MovingMNIST. The top half are two MovingMNIST sequences, where their initial conditions only slightly differ in the the initial velocities. The bottom half are two \nbody{} sequences. \nbody{} sequence 1 has exactly the same initial condition as MovingMNIST sequence 1. \nbody{} sequence 2 has exactly the same initial condition as MovingMNIST sequence 2. The final positions of digits in MovingMNIST after $20$ steps evolution only slightly differ from each other, while the differences are much more significant in the final frames of \nbody{} sequences.}
    \label{fig:chaos_nbody}
\end{figure}

The Earth system is chaotic~\cite{tapley2004grace}, meaning that the future is very sensitive to the initial conditions. In~\figref{fig:chaos_nbody}, we illustrate the chaotic effect of \nbody{}. By only slightly changing the initial velocities of the digits, the positions of the digits after $20$ steps change significantly. \nbody{} is thus much more challenging than MovingMNIST.

\section{Evaluation Metrics in SEVIR}
In~\secref{sec:exp_sevir} and~\tabref{table:sevir_csi}, we follow~\cite{veillette2020sevir} to use the $\mathtt{Critical\ Success\ Index}$ ($\mathtt{CSI}$) for prediction quality evaluation\footnote{The implementation details of $\mathtt{CSI}$ used in~\secref{sec:exp_sevir} and~\tabref{table:sevir_csi} follow \url{https://github.com/MIT-AI-Accelerator/neurips-2020-sevir}}.
Besides~\cite{veillette2020sevir}, the SEVIR team holds the \emph{SEVIR Dataset Challenge}\footnote{Challenge website at \url{https://sevir.mit.edu/nowcasting}}, where the data involved are exactly the same while the evaluation metrics are slightly different\footnote{The implementation details of $\mathtt{CSI}$ used in the SEVIR Dataset Challenge are available at \url{https://github.com/MIT-AI-Accelerator/sevir_challenges}.}.

Specifically, \cite{veillette2020sevir} used six precipitation thresholds which correspond to pixel values $[219, 181, 160, 133, 74, 16]$. 
The prediction and the ground-truth are rescaled back to the range 0-255.
The $\mathtt{\#Hits}(\tau)$ (truth$\geq\tau$, pred$\geq\tau$), $\mathtt{\#Misses}(\tau)$ (truth$\geq\tau$, pred$<\tau$) and $\mathtt{\#F.Alarms}(\tau)$ (truth$<\tau$, pred$\geq\tau$) at certain threshold $\tau$ are counted over all test pixels as shown in~\eqnref{eq:hits_misses_fas}.

\begin{align}
    \centering
    \begin{aligned}
        \mathtt{\#Hits}(\tau) &= \sum^N_{n=1}\sum^T_{t=1}\sum^H_{h=1}\sum^W_{w=1}\mathtt{Hits}(\tau)[n, t , h, w] \\
        \mathtt{\#Misses}(\tau) &= \sum^N_{n=1}\sum^T_{t=1}\sum^H_{h=1}\sum^W_{w=1}\mathtt{Misses}(\tau)[n, t , h, w] \\
        \mathtt{\#F.Alarms}(\tau) &= \sum^N_{n=1}\sum^T_{t=1}\sum^H_{h=1}\sum^W_{w=1}\mathtt{F.Alarms}(\tau)[n, t , h, w] \\
        \mathtt{CSI}\mbox{-}\tau &= \frac{\mathtt{\#Hits}(\tau)}{\mathtt{\#Hits}(\tau)+\mathtt{\#Misses}(\tau)+\mathtt{\#F.Alarms}(\tau)} \\
        \mathtt{CSI}\mbox{-}\mathtt{M} &= \frac{1}{6}\sum_{\tau\in[219, 181, 160, 133, 74, 16]}\mathtt{CSI}\mbox{-}\tau
    \end{aligned}
    \label{eq:hits_misses_fas}
\end{align}

$\tau\in[219, 181, 160, 133, 74, 16]$ is one of the thresholds. 
$\mathtt{Hits}(\tau),\mathtt{Misses}(\tau),\mathtt{F.Alarms}(\tau)\in[0,1]^{N\times T\times H\times W}$, where $N$ is the test dataset size, $T$ is the forecasting horizon, $H$ and $W$ are the height and width, respectively.
We denote the average $\mathtt{CSI}\mbox{-}\tau$ over the thresholds $[219, 181, 160, 133, 74, 16]$ as $\mathtt{CSI}\mbox{-}\mathtt{M}$.
The results using the above evaluation metrics are demonstrated in~\secref{sec:exp_sevir} and~\tabref{table:sevir_csi}.

The calculation of $\mathtt{CSI}$ is slightly different in the SEVIR Dataset Challenge, as shown in~\eqnref{eq:metrics_challenge}.

\begin{align}
    \centering
    \begin{aligned}
        \mathtt{\#Hits}(\tau, t) &= \sum^N_{n=1}\sum^H_{h=1}\sum^W_{w=1}\mathtt{Hits}(\tau)[n, t , h, w] \\
        \mathtt{\#Misses}(\tau, t) &= \sum^N_{n=1}\sum^H_{h=1}\sum^W_{w=1}\mathtt{Misses}(\tau)[n, t , h, w] \\
        \mathtt{\#F.Alarms}(\tau, t) &= \sum^N_{n=1}\sum^H_{h=1}\sum^W_{w=1}\mathtt{F.Alarms}(\tau)[n, t , h, w] \\
        \mathtt{CSI}\mbox{-}\tau(t) &= \frac{\mathtt{\#Hits}(\tau, t)}{\mathtt{\#Hits}(\tau, t)+\mathtt{\#Misses}(\tau, t)+\mathtt{\#F.Alarms}(\tau, t)} \\
        \mathtt{CSI}\mbox{-}\tau &= \frac{1}{T}\sum_t^T\mathtt{CSI}\mbox{-}\tau(t) \\
        \mathtt{CSI}\mbox{-}\mathtt{M}3 &= \frac{1}{3}\sum_{\tau\in[133, 74, 16]}\mathtt{CSI}\mbox{-}\tau \\
        \mathtt{CSI}\mbox{-}\mathtt{M}6 &= \frac{1}{6}\sum_{\tau\in[219, 181, 160, 133, 74, 16]}\mathtt{CSI}\mbox{-}\tau \\
    \end{aligned}
    \label{eq:metrics_challenge}
\end{align}

The $\mathtt{CSI}\mbox{-}\tau$, at certain threshold, is the mean of $\mathtt{CSI}\mbox{-}\tau(t)$ over forecasting horizon $T$.
We denote the average $\mathtt{CSI}\mbox{-}\tau$ over the three thresholds $[133, 74, 16]$ used in the SEVIR Dataset Challenge as $\mathtt{CSI}\mbox{-}\mathtt{M}3$, and the average $\mathtt{CSI}\mbox{-}\tau$ over the six thresholds $[219, 181, 160, 133, 74, 16]$ used in~\cite{veillette2020sevir} as $\mathtt{CSI}\mbox{-}\mathtt{M}6$.
The experiment results evaluated using the metrics in the SEVIR Dataset Challenge are listed in~\tabref{table:sevir_csi_challenge}.
Earthformer still consistently outperforms the baselines in almost all the metrics.

\begin{table}[!tb]
\caption{
Performance comparison on SEVIR using metrics in SEVIR Dataset Challenge. 
The $\mathtt{CSI}$ is calculated at precipitation thresholds $[219, 181, 160, 133, 74, 16]$. 
Different from~\tabref{table:sevir_csi}, $\mathtt{CSI}\mbox{-}\tau = \frac{1}{T}\sum_t^T\mathtt{CSI}\mbox{-}\tau(t)$ is the mean of $\mathtt{CSI}\mbox{-}\tau(t)$ over forecasting horizon $T$.
$\mathtt{CSI}\mbox{-}\mathtt{M}3$ and $\mathtt{CSI}\mbox{-}\mathtt{M}6$ are the average of $\mathtt{CSI}\mbox{-}\tau$ over thresholds $[133, 74, 16]$ and $[219, 181, 160, 133, 74, 16]$, respectively.}
\label{table:sevir_csi_challenge}
	\begin{center}
	\resizebox{0.99\textwidth}{!}{
	\begin{tabular}{l|c|c|cccccc|cc|cc}
	\toprule[1.5pt]
	\multirow{2}{*}{Model}& \multirow{2}{*}{\#Param. (M)}& \multirow{2}{*}{GFLOPS} & \multicolumn{10}{|c}{Metrics}\\
 						                &           &       & $\mathtt{CSI}\mbox{-}219$ $\uparrow$ & $\mathtt{CSI}\mbox{-}181$ $\uparrow$ & $\mathtt{CSI}\mbox{-}160$ $\uparrow$ & $\mathtt{CSI}\mbox{-}133$ $\uparrow$ & $\mathtt{CSI}\mbox{-}74$ $\uparrow$ & $\mathtt{CSI}\mbox{-}16$ $\uparrow$ & $\mathtt{CSI}\mbox{-}\mathtt{M}3$ $\uparrow$  & $\mathtt{CSI}\mbox{-}\mathtt{M}6$ $\uparrow$ & MSE ($10^{-3}$) $\downarrow$    & MAE ($10^{-2}$) $\downarrow$\\
	\midrule\midrule
	Persistence			                & -                 & -     			& 0.0575				& 0.1056				& 0.1374				& 0.2259				& 0.4796				& 0.6109				& 0.4386                & 0.2695				& 11.5283				& 4.4349				\\
	UNet~\cite{veillette2020sevir}      &\ \ 16.6           &\ \ 33	            & 0.0521				& 0.1364				& 0.1887				& 0.3032				& 0.6542				& 0.7444				& 0.5673                & 0.3465				& 4.1119				& 2.7633				\\
	ConvLSTM~\cite{shi2015convolutional}&\ \ 14.0           & 527	            & 0.1071				& 0.2275				& 0.2829				& 0.4155				& 0.6873				& 0.7555				& 0.6194                & 0.4126				& 3.7532				& 2.5898				\\
	PredRNN~\cite{wang2022predrnn}      &\ \ 46.6           & 328				& 0.1164				& 0.2246				& 0.2718				& 0.3873				& 0.6744				& 0.7544				& 0.6054                & 0.4048				& 3.9014				& 2.6963				\\
	PhyDNet~\cite{guen2020disentangling}&\ \ 13.7           & 701				& 0.1041				& 0.2123				& 0.2583				& 0.3755				& 0.6627				& 0.7176				& 0.5853                & 0.3884				& 4.8165				& 3.1896				\\
	E3D-LSTM~\cite{wang2018eidetic}     &\ \ 35.6           & 523			    & 0.1125				& 0.2218				& 0.2637				& 0.3847				& 0.6674				& 0.7591				& 0.6037                & 0.4015				& 4.1702				& \textbf{2.5023}		\\
	Rainformer~\cite{bai2022rainformer} & 184.0             & 170				& 0.0745				& 0.1580				& 0.2087				& 0.3397				& 0.6599				& 0.7308				& 0.5768                & 0.3619				& 4.0272				& 3.0711				\\
	\midrule
	Earthformer w/o global              &\ \ 13.1           & 257				& \underline{0.1429}	& \underline{0.2643}	& \underline{0.3086}	& \underline{0.4215}	& \underline{0.6885}	& \underline{0.7671}	& \underline{0.6257}    & \underline{0.4321}	& \underline{3.7006}	& 2.5306				\\
	Earthformer			                &\ \ 15.1           & 257				& \textbf{0.1480}		& \textbf{0.2748}		& \textbf{0.3126}		& \textbf{0.4231}		& \textbf{0.6886}		& \textbf{0.7682}		& \textbf{0.6266}       & \textbf{0.4359}		& \textbf{3.6702}		& \underline{2.5112}	\\
	\bottomrule[1.5pt]
	\end{tabular}
	}
	\end{center}
\end{table}

\end{document}